%% file: main.tex
\documentclass[journal]{IEEEtran}
\ifCLASSINFOpdf
\else
\fi
\usepackage[table,xcdraw]{xcolor}
\usepackage{amssymb}

\definecolor{rblue}{rgb}{0,0.5,1}
\definecolor{awesome}{rgb}{1.0, 0.13, 0.32}
\definecolor{hollywoodcerise}{rgb}{0.96, 0.0, 0.63}
\definecolor{lasallegreen}{rgb}{0.03, 0.47, 0.19}
\definecolor{hanpurple}{rgb}{0.32, 0.09, 0.98}
\definecolor{green(pigment)}{rgb}{0.0, 0.65, 0.31}

\usepackage[pagebackref=false,breaklinks=true,colorlinks,bookmarks=false]{hyperref}
\hypersetup{colorlinks=true,linkcolor={red},citecolor={hanpurple},urlcolor={magenta}}
\usepackage{times}
\usepackage{epsfig}
\usepackage{graphicx}
\usepackage{amsmath}
\usepackage{amssymb}
\usepackage{array}
\usepackage{float}
\usepackage{placeins}
\usepackage{morewrites}
\usepackage{makecell}

\usepackage{caption}
\usepackage{subcaption}
\usepackage{tabularx}
\usepackage{rotating}
\usepackage{verbatim}
\usepackage{etoolbox}
\usepackage[normalem]{ulem}
\usepackage{booktabs}
\usepackage{multirow}
\usepackage{lipsum}
\usepackage{stmaryrd}
\usepackage{stackengine}
\usepackage{makecell}
\usepackage{pifont}
\usepackage{cancel}
\usepackage{adjustbox}
\usepackage{dblfloatfix}
\usepackage[table]{xcolor}

\newcommand{\camera}{\includegraphics[width=3.8mm]{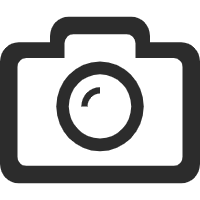}}
\newcommand{\LiDAR}{\includegraphics[width=4mm]{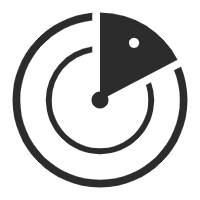}}
\newcommand{\event}{\includegraphics[width=4mm]{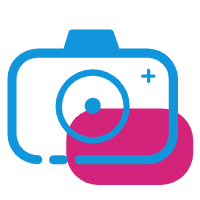}}
\newcommand{\radar}{\includegraphics[width=4mm]{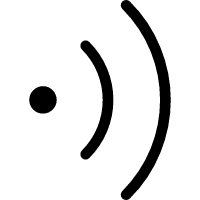}}
\newcommand{\tick}{\includegraphics[width=4mm]{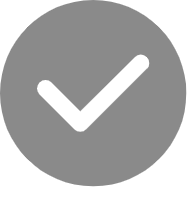}}
\newcommand{\cross}{\includegraphics[width=4mm]{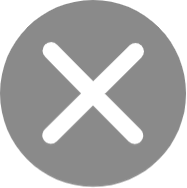}}
\newcommand{\cir}{\includegraphics[width=4mm]{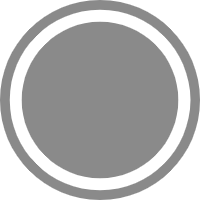}}

\newcommand{\real}{\includegraphics[width=4mm]{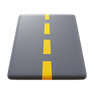}}
\newcommand{\syn}{\includegraphics[width=4mm]{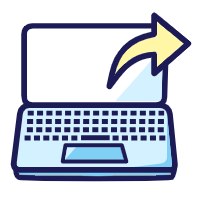}}

\usepackage[capitalize]{cleveref}
\crefname{section}{Sec.}{Secs.}
\Crefname{section}{Section}{Sections}
\Crefname{table}{Table}{Tables}
\crefname{table}{Tab.}{Tabs.}
\usepackage{colortbl}

\definecolor{car_s}{rgb}{0.39215686, 0.58823529, 0.96078431}
\definecolor{bicycle_s}{rgb}{0.39215686, 0.90196078, 0.96078431}
\definecolor{motorcycle_s}{rgb}{0.11764706, 0.23529412, 0.58823529}
\definecolor{truck_s}{rgb}{0.31372549, 0.11764706, 0.70588235}
\definecolor{other-vehicle_s}{rgb}{0.39215686, 0.31372549, 0.98039216}
\definecolor{person_s}{rgb}{1.        , 0.11764706, 0.11764706}
\definecolor{bicyclist_s}{rgb}{1.        , 0.15686275, 0.78431373}
\definecolor{motorcyclist_s}{rgb}{0.58823529, 0.11764706, 0.35294118}
\definecolor{road_s}{rgb}{1.        , 0.        , 1.        }
\definecolor{parking_s}{rgb}{1.        , 0.58823529, 1.        }
\definecolor{sidewalk_s}{rgb}{0.29411765, 0.        , 0.29411765}
\definecolor{other-ground_s}{rgb}{0.68627451, 0.        , 0.29411765}
\definecolor{building_s}{rgb}{1.        , 0.78431373, 0.        }
\definecolor{fence_s}{rgb}{1.        , 0.47058824, 0.19607843}
\definecolor{vegetation_s}{rgb}{0.        , 0.68627451, 0.        }
\definecolor{trunk_s}{rgb}{0.52941176, 0.23529412, 0.        }
\definecolor{terrain_s}{rgb}{0.58823529, 0.94117647, 0.31372549}
\definecolor{pole_s}{rgb}{1.        , 0.94117647, 0.58823529}
\definecolor{traffic-sign_s}{rgb}{1.        , 0.        , 0.    }

\definecolor{car_d}{rgb}{0.0, 0.0, 0.55686275}
\definecolor{bicycle_d}{rgb}{0.46666667, 0.04313725, 0.12549020}
\definecolor{motorcycle_d}{rgb}{0.0, 0.0, 0.90196078}
\definecolor{truck_d}{rgb}{0.0, 0.0, 0.27450980}
\definecolor{other-vehicle_d}{rgb}{0.90196078, 0.58823529, 0.54901961}
\definecolor{person_d}{rgb}{0.86274510, 0.07843137, 0.23529412}
\definecolor{road_d}{rgb}{0.50196078, 0.25098039, 0.50196078}
\definecolor{sidewalk_d}{rgb}{0.95686275, 0.13725490, 0.90980392}
\definecolor{building_d}{rgb}{0.27450980, 0.27450980, 0.27450980}
\definecolor{fence_d}{rgb}{0.74509804, 0.60000000, 0.60000000}
\definecolor{vegetation_d}{rgb}{0.41960784, 0.55686275, 0.13725490}
\definecolor{terrain_d}{rgb}{0.59607843, 0.98431373, 0.59607843}
\definecolor{pole_d}{rgb}{0.60000000, 0.60000000, 0.60000000}
\definecolor{traffic-sign_d}{rgb}{0.86274510, 0.86274510, 0.0}
\makeatletter
\newcommand{\car@semkitfreq}{3.92}
\newcommand{\bicycle@semkitfreq}{0.03}
\newcommand{\motorcycle@semkitfreq}{0.03}
\newcommand{\truck@semkitfreq}{0.16}
\newcommand{\othervehicle@semkitfreq}{0.20}
\newcommand{\person@semkitfreq}{0.07}
\newcommand{\bicyclist@semkitfreq}{0.07}
\newcommand{\motorcyclist@semkitfreq}{0.05}
\newcommand{\road@semkitfreq}{15.30}  %
\newcommand{\parking@semkitfreq}{1.12}
\newcommand{\sidewalk@semkitfreq}{11.13}  %
\newcommand{\otherground@semkitfreq}{0.56}
\newcommand{\building@semkitfreq}{14.1}  %
\newcommand{\fence@semkitfreq}{3.90}
\newcommand{\vegetation@semkitfreq}{39.3}  %
\newcommand{\trunk@semkitfreq}{0.51}
\newcommand{\terrain@semkitfreq}{9.17} %
\newcommand{\pole@semkitfreq}{0.29}
\newcommand{\trafficsign@semkitfreq}{0.08}
\newcommand{\semkitfreq}[1]{{\csname #1@semkitfreq\endcsname}}
\makeatletter
\newcommand{\car@dsecfreq}{1.71}
\newcommand{\bicycle@dsecfreq}{0.02}
\newcommand{\motorcycle@dsecfreq}{0.01}
\newcommand{\truck@dsecfreq}{0.17}
\newcommand{\othervehicle@dsecfreq}{0.26}
\newcommand{\person@dsecfreq}{0.05}
\newcommand{\road@dsecfreq}{29.02}
\newcommand{\sidewalk@dsecfreq}{16.58}
\newcommand{\building@dsecfreq}{17.22}
\newcommand{\fence@dsecfreq}{5.96}
\newcommand{\vegetation@dsecfreq}{26.88}
\newcommand{\terrain@dsecfreq}{1.50}
\newcommand{\pole@dsecfreq}{0.54}
\newcommand{\trafficsign@dsecfreq}{0.09}
\newcommand{\dsecfreq}[1]{{\csname #1@dsecfreq\endcsname}}
\makeatother

\begin{document}
%
\title{Event-aided Semantic Scene Completion}
%
%
%

\author{Shangwei Guo$^{1,*}$, Hao Shi$^{1,*}$, Song Wang$^{3}$, Xiaoting Yin$^{1}$, Kailun Yang$^{2,\dag}$, and Kaiwei Wang$^{1,\dag}$
\thanks{This work was supported in part by Zhejiang Provincial Natural Science Foundation of China (Grant No. LZ24F050003), the National Natural Science Foundation of China (Grant No. 12174341 and No. 62473139), and in part by Shanghai SUPREMIND Technology Company Ltd.
}
\thanks{$^{1}$The authors are with the State Key Laboratory of Extreme Photonics and Instrumentation, Zhejiang University, Hangzhou 310027, China (email: wangkaiwei@zju.edu.cn).}%
\thanks{$^{2}$The author is with the School of Robotics and the National Engineering Research Center of Robot Visual Perception and Control Technology, Hunan University, Changsha 410082, China (email: kailun.yang@hnu.edu.cn).}%
\thanks{$^{3}$The author is with the College of Computer Science and Technology, Zhejiang University, Hangzhou 310027, China.}%
\thanks{$^*$Equal contribution.}%
\thanks{$^\dag$Corresponding authors: Kaiwei Wang and Kailun Yang.}%
}


%



\maketitle

\begin{abstract}
\input{Tex_content/abstract}
\end{abstract}


%
\IEEEpeerreviewmaketitle

\input{Tex_content/intro}

\input{Tex_content/related_work}

%
%


\input{Tex_content/benchmark}
\input{Tex_content/method}
\input{Tex_content/Exp}

\input{Tex_content/conclusion}

{\small
\bibliographystyle{IEEEtran}
\bibliography{bib}
}

\end{document}

%% file: Tex_content/abstract.tex
Autonomous driving systems rely on robust 3D scene understanding. Recent advances in Semantic Scene Completion (SSC) for autonomous driving underscore the limitations of RGB-based approaches, which struggle under motion blur, poor lighting, and adverse weather. Event cameras, offering high dynamic range and low latency, address these challenges by providing asynchronous data that complements RGB inputs. We present DSEC-SSC, the first real-world benchmark specifically designed for event-aided SSC, which includes a novel 4D labeling pipeline for generating dense, visibility-aware labels that adapt dynamically to object motion. Our proposed RGB-Event fusion framework, EvSSC, introduces an Event-aided Lifting Module (ELM) that effectively bridges 2D RGB-Event features to 3D space, enhancing view transformation and the robustness of 3D volume construction across SSC models. Extensive experiments on DSEC-SSC and simulated SemanticKITTI-E demonstrate that EvSSC is adaptable to both transformer-based and LSS-based SSC architectures. Notably, evaluations on SemanticKITTI-C demonstrate that EvSSC achieves consistently improved prediction accuracy across five degradation modes and both In-domain and Out-of-domain settings, achieving up to a $52.5\%$ relative improvement in mIoU when the image sensor partially fails. Additionally, we quantitatively and qualitatively validate the superiority of EvSSC under motion blur and extreme weather conditions, where autonomous driving is challenged. The established datasets and our codebase will be made publicly at \url{https://github.com/Pandapan01/EvSSC}.

%% file: Tex_content/intro.tex
\section{Introduction}
\IEEEPARstart{S}{emantic} Scene Completion (SSC), or 3D semantic occupancy prediction, is critical in autonomous driving and robotics, where understanding complex environments is essential for safe navigation~\cite{zhang2024vision3D}. 
The task involves generating a dense scene representation encompassing both geometric and semantic details~\cite{xu2023survey, xu2025survey}.

\begin{figure}[t]
\centering
\includegraphics[width=1.0\linewidth]{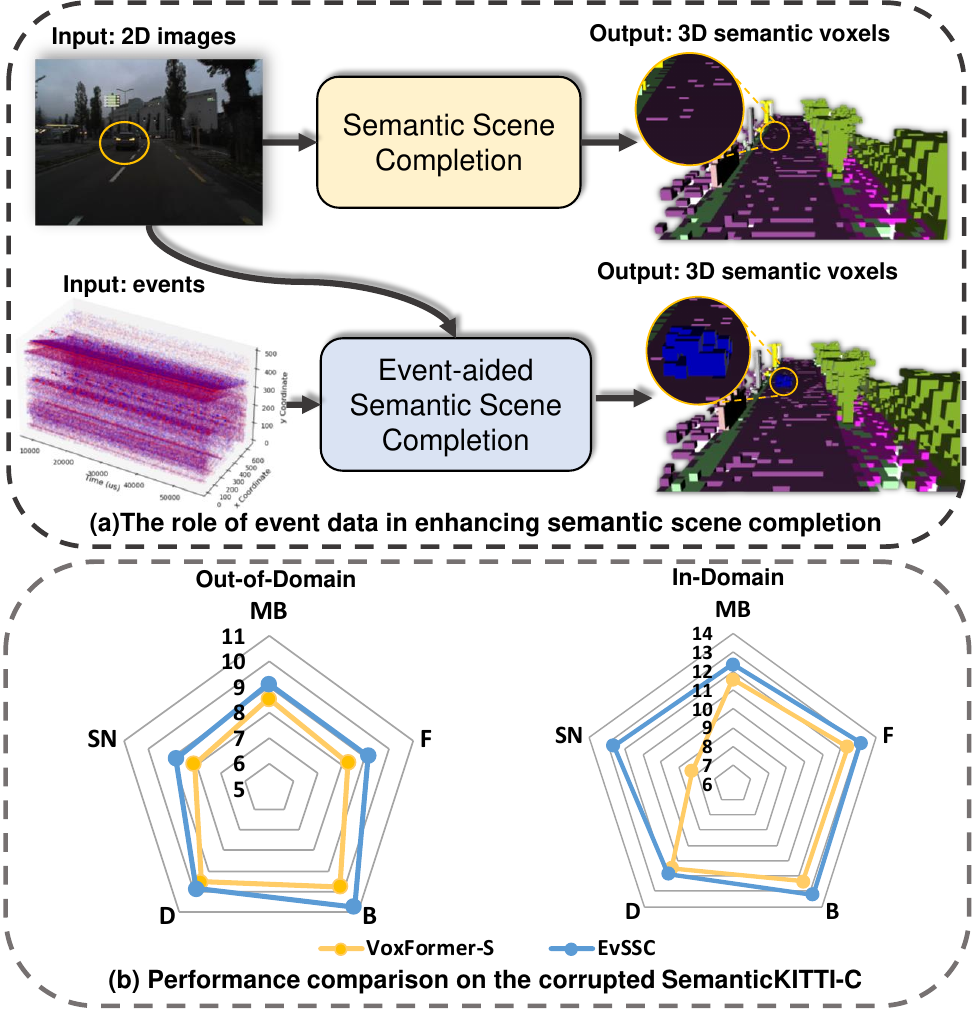}
\caption{(a) \textbf{Role of event data in enhancing semantic scene completion:} Under challenging lighting conditions, RGB-based methods struggle to detect low-contrast objects, whereas event data enhances visibility and improves 3D occupancy predictions. 
(b) \textbf{Performance comparison on the corrupted SemanticKITTI-C:} mIoU results across Out-of-Domain and In-Domain scenarios, with and without event data integration, showing scores under various conditions, including Motion Blur (MB), Fog (F), Brightness (B), Darkness (D), and Shot Noise (SN).}
\label{fig:core_concepts}
\end{figure}

However, traditional camera-based SSC methods often struggle in challenging lighting and weather conditions due to issues such as motion blur, low-light scenarios~\cite{Gehrig2024LowLatency}, and adverse weather effects~\cite{Kong2023Robo3D}. Event cameras offer a promising visual solution, providing high-temporal-resolution data with advantages like robustness to motion blur and low-latency response~\cite{gallego2020eventbased}. As shown in Fig.~\ref{fig:core_concepts}(a), in low-light scenarios, event data captures low-contrast objects, such as vehicles, that RGB fails to distinguish. These features make event cameras highly suitable for tasks requiring consistent and reliable scene understanding, particularly in degraded visual environments. 
The complementary nature of event and RGB data allows event cameras to capture rapid motion dynamics that are often missed by standard cameras, whereas RGB images contribute spatial detail. This bidirectional complementarity enhances 3D occupancy prediction, making autonomous driving safer and more responsive.

\input{tables/Existing_Datasets}

Despite recent advancements and the establishment of benchmarks~\cite{behley2019semantickitti,pan2020semanticposs,tian2024occ3d,wang2023openoccupancy,tong2023scene_as_occupancy},
event-based semantic scene completion remains underexplored. There is still a lack of event-modality SSC datasets due to the challenges in creating accurate labels. 
To address this dearth, we introduce \emph{DSEC-SSC} (see Tab.~\ref{table:dataset}), the first real-world event-camera dataset for event-aided semantic scene completion, featuring dedicated processing to ensure reliable ground truth for dynamic objects. 
We also propose a semi-automatic label generation pipeline, which effectively bridges the domain gap between datasets to handle dynamic objects, even in the absence of 2D and 3D ground truth annotations for object detection during the establishment of our DSEC-SSC. 
The semi-automated annotation pipeline (Fig.~\ref{fig:dsec_ssc_pipeline}) simplifies complex point cloud labeling into efficient 2D image-based annotation and enables precise 4D reconstruction of dynamic objects, producing high-accuracy spatiotemporal ground truth. The method is sensor-agnostic, and compatible with various LiDAR brands and models, offering a practical approach to generating accurate 3D/4D occupancy data. 
Moreover, we benchmark a range of classic and recent SSC models on the DSEC-SSC dataset.

Further, we propose \emph{EvSSC}, an event-aided framework for Semantic Scene Completion (SSC) that enhances 3D occupancy prediction by integrating event data. 
Central to EvSSC is the \emph{Event-aided Lifting Module (ELM)}, which performs the crucial lifting process, transforming 2D features into a coherent 3D space. 
This transformation is essential in 2D-to-3D occupancy prediction, as it aligns 2D features with the 3D spatial structure of the scene, directly impacting accuracy and stability.
To effectively fuse image and event data, we explore three paradigms (Fig.~\ref{fig:fusion&framework}(a)): fusion-then-lifting, decode-then-fusion, and our preferred fusion-based lifting. Fusion-then-lifting enhances early feature alignment but risks spatial inconsistencies, while decode-then-fusion maintains modality independence but delays interaction. 
Fusion-based lifting integrates event data directly in the lifting stage, preserving spatial fidelity and capturing temporal dynamics.
The ELM in EvSSC uses multi-scale attention during lifting to adaptively merge 2D event and RGB features into 3D space, resulting in robust SSC predictions in dynamic environments.

To verify the effectiveness of our approach, we benchmark EvSSC on the newly introduced DSEC-SSC and the simulated SemanticKITTI-E datasets. 
EvSSC consistently outperforms baseline models, demonstrating significant gains across both transformer-based~\cite{li2023voxformer} and LSS-based~\cite{mei2024sgn} SSC architectures. 
These results underscore the advantages of our fusion-based lifting method in leveraging event data for more accurate 3D occupancy predictions.
Furthermore, to evaluate EvSSC’s robustness in challenging conditions, we conducted tests across five common degradation scenarios in autonomous driving: motion blur, fog, low and high brightness, and noise. 
As shown in Fig.~\ref{fig:core_concepts}(b), EvSSC achieves up to a $52.5\%$ relative improvement in mIoU on the corrupted SemanticKITTI-C dataset, confirming its resilience in degraded environments.

In summary, we deliver the following contributions:
\begin{itemize}
    \item We establish \emph{DSEC-SSC}, the first real-world event camera dataset tailored for semantic scene completion, and benchmark a variety of classic and recent SSC models.
    \item We propose \emph{EvSSC}, with \emph{Event-aided Lifting Module (ELM)}, achieving high accuracy in semantic scene completion, as validated through extensive benchmarks on the \emph{DSEC-SSC} and \emph{SemanticKITTI-E} datasets.
    \item We perform comprehensive experiments across diverse corruption scenarios, demonstrating the robustness and superiority of our event-based approach in more challenging conditions.
\end{itemize}

%% file: tables/Existing_Datasets.tex
\begin{table*}[]
\caption{Comparison of datasets for semantic scene completion. Abbreviations: \real~(Real-World Data), \syn~(Synthetic Data), \camera~(Camera), \radar~(Radar), \LiDAR~(LiDAR), \event~(Event Camera), \cir~(Irrelevant).}
\centering
\begin{adjustbox}{width=\textwidth}
\begin{tabular}{lccccccc}
\hline  
\textbf{Datasets} & \textbf{Year} & \textbf{Real-World}
& \textbf{Modality} & \textbf{\#Classes} & \textbf{Voxel Size} & \textbf{Dynamic Object Processing}\\ 
\hline  
\hline
SemanticKITTI~\cite{behley2019semantickitti} & 2019 & \real & \camera\LiDAR &
28 & 256$\times$256$\times$32 & \cross\\
SemanticPOSS~\cite{pan2020semanticposs}& 2020 & \real & \LiDAR &
14 & - & \cross \\
Occ3D-nuScenes~\cite{tian2024occ3d} & 2023 & \real  & \camera &
16 & 200$\times$200$\times$16 & \tick\\
Occ3D-Waymo~\cite{tian2024occ3d} & 2023 & \syn & \camera &
14 & 200$\times$200$\times$32 & \cir \\
nuScenes-Occupancy~\cite{wang2023openoccupancy} & 2023 & \real  & \camera\LiDAR &
16 & 512$\times$512$\times$40 & \tick\\
OpenOcc~\cite{tong2023scene_as_occupancy} & 2023 & \real & \camera\LiDAR &
16 & 200$\times$200$\times$16 &\tick \\
SSCBench-KITTI360~\cite{li2023sscbench} & 2024 & \real & \camera\LiDAR & 19 & 256$\times$256$\times$32 & \tick\\ 
nuCraft~\cite{zhu2024nucraft} &2024 & \real & \camera\LiDAR & 16 & 1024$\times$1024$\times$80 & \tick\\
PointSSC~\cite{yan2024pointssc}&2024 & \real & \camera\LiDAR & 9 & - & \tick \\ 
WildOcc~\cite{zhai2024wildocc} & 2024 & \real & \camera\LiDAR & 20 & 256$\times$256$\times$32 & \cross\\ 
V2VSSC~\cite{zhang2024v2vssc}& 2024 & \syn & \camera\LiDAR & 6 & 128$\times$128$\times$20 & \cir \\
ScribbleSC~\cite{wang2024label_efficient} & 2024  &\real & \camera\LiDAR & 19 & 256$\times$256$\times$32 & \cross \\
K-Radar~\cite{ding2024radarocc,paek2022k-radar} & 2024  &\real & \camera\LiDAR\radar & 2 & - & \tick \\
OmniHD-Scenes~\cite{zheng2024omnihd} & 2024 &\real &\camera\LiDAR\radar & 11 & - & \tick \\ 
\rowcolor{gray!20}
SemanticKITTI-E (Ours) & 2025 & \syn & \event\camera\LiDAR & 28 & 256$\times$256$\times$32 & \cross\\
\rowcolor{gray!20}
DSEC-SSC (Ours) & 2025 & \real & \event\camera\LiDAR & 14 & 128$\times$128$\times$16 & \tick\\

\hline
\end{tabular} 
\end{adjustbox}
\label{table:dataset}
\end{table*}

%% file: Tex_content/related_work.tex
\section{Related Work}

\begin{figure*}[!t]
\centering
\includegraphics[width=1.0\linewidth]{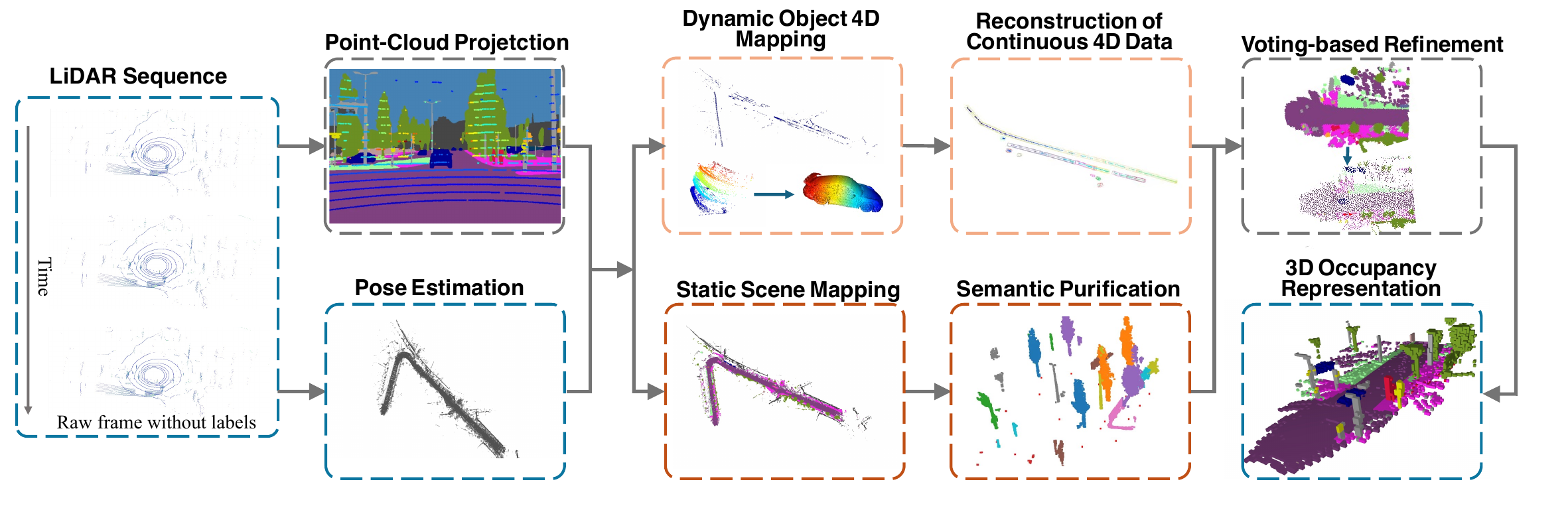}
\caption{\textbf{Overview of the occupancy label generation pipeline.} The pipeline consists of three main steps: Semantic-Maps-Guided Dynamic Object Segmentation (Sec.~\ref{sec:semantic_guided_dynamic_object_segmentation}), Dynamic Object 4D Reconstruction (Sec.~\ref{sec:dynamic_objects_4d_reconstruction}), and Probability-Guided Voxel Refinement (Sec.~\ref{sec:probability_guided_voxel_refinement}).  
}
\label{fig:dsec_ssc_pipeline}
\end{figure*}

\subsection{Semantic Scene Completion}
Existing Semantic Scene Completion (SSC) approaches can be mainly classified into LiDAR-based, camera-based, and modality-fusion methods.
Although LiDAR-based~\cite{xia2023scpnet,cheng2021s3cnet,yan2021js3c_net,roldao2020lmscnet,jang2024talos,cao2024slcf,peng2022mass} and modality-fusion methods~\cite{pan2024co_occ,ming2024occfusion,lu2024_ipvoxelnet,xue2024bi_ssc,xue2024bi_ssc,liu20242d_guided,ding2024radarocc,zhang2024radocc,cui2024loma} can deliver relatively strong performance, camera-based methods are often preferred for practical deployment due to their lower economic costs and superior real-time capabilities.
MonoScene~\cite{cao2022monoscene} introduces the first approach to infer 3D SSC from a single monocular image. 
Recent advancements in SSC have developed from dual-path transformer model~\cite{zhang2023occformer} to two-stage architectures with masked autoencoders~\cite{li2023voxformer}, followed by the introduction of context-aware instance queries~\cite{jiang2024symphonize} and hybrid guidance to improve feature separation~\cite{mei2024sgn,shi2024panossc}.
Further research expands into several aspects with efficiency optimization~\cite{wang2024not,hou2024fastocc,zhao2024lowrankocc,wang2024reliocc,yang2024adaptiveocc}, 
offboard perception~\cite{shi2024occfiner,ma2024zopp,zheng2024occworld,ouyang2024linkocc},
3D representation~\cite{huang2024gaussianformer,huang2023tpvformer,shi2024occupancy_set_points,xiao2024instance_aware,wang2024opus}, 
sparse feature processing~\cite{tang2024sparseocc,liu2024fully_sparse,wei2023surroundocc}, 
open-vocabulary recognition~\cite{vobecky2024pop_3d,zheng2024veon}, 
self-supervision~\cite{huang2024selfocc,hayler2024s4c}, 
and context enhancement~\cite{yu2024cgformer,zhao2024hybridocc,ma2024cotr,li2024htcl,li2024viewformer},

While conventional cameras are cost-effective, their perception capabilities are highly susceptible to poor weather and illumination conditions.
Event cameras provide robust, high-frequency information in complex environments. 
Therefore, we introduce a real-world 3D occupancy prediction dataset, \emph{DSEC-SSC}, incorporating event modality and for the first time accomplishing event-aided SSC.

\subsection{Event-driven Semantic Dense Understanding}
Recently, event cameras have gained prominence in semantic dense understanding due to their high dynamic range, low latency, and low bandwidth. 
RGB-Event fusion~\cite{xia2023cmda,zhang2023cmx,zhen2025event_guided,xie2024eisnet} significantly benefits the performance of vision tasks. 
Many studies have explored image-event fusion, fully leveraging texture features of RGB and high-frequency characteristics of events.
Event-based semantic segmentation, first introduced in Ev-SegNet~\cite{alonso2019ev_segnet}, capitalizes on asynchronous events to achieve significant advances.
Subsequent works focus on scene-adaptive rendering~\cite{low2021superevents}, unsupervised adaptation~\cite{sun2022ess}, event-specific attention~\cite{jia2023event_posterior}, bidirectional fusion~\cite{zhang2021issafe,zhang2021exploring}, and multi-branch feature extraction~\cite{zhang2024spikingedn,long2024spike_brgnet}, advancing the robustness and flexibility of event representation. 
Event cameras have also supported advancements in object detection~\cite{cao2024embracing,liu2024enhancing_traffic,li2023sodformer}, deblurring~\cite{sun2022event_deblurring,kim2024frequency,yang2024latency}, flow estimation~\cite{wan2023rpeflow,ye2023towards_event}, and tracking~\cite{zhang2023frame_event_tracking,wang2023visevent,wang2024towards_robust_tracking}. 
Yet, event-aided SSC remains largely underexplored. 
We propose \emph{EvSSC}, containing our custom-designed \emph{Event-aided Lifting Module (ELM)} for 2D-to-3D view transformation, tailored for the SSC task, enabling effective cross-modal fusion.

%% file: Tex_content/benchmark.tex
%
\section{DSEC-SSC: an Event-based SSC Dataset}
\label{sec:benchmark}

\begin{figure}[!t]
\includegraphics[width=1.0\linewidth]{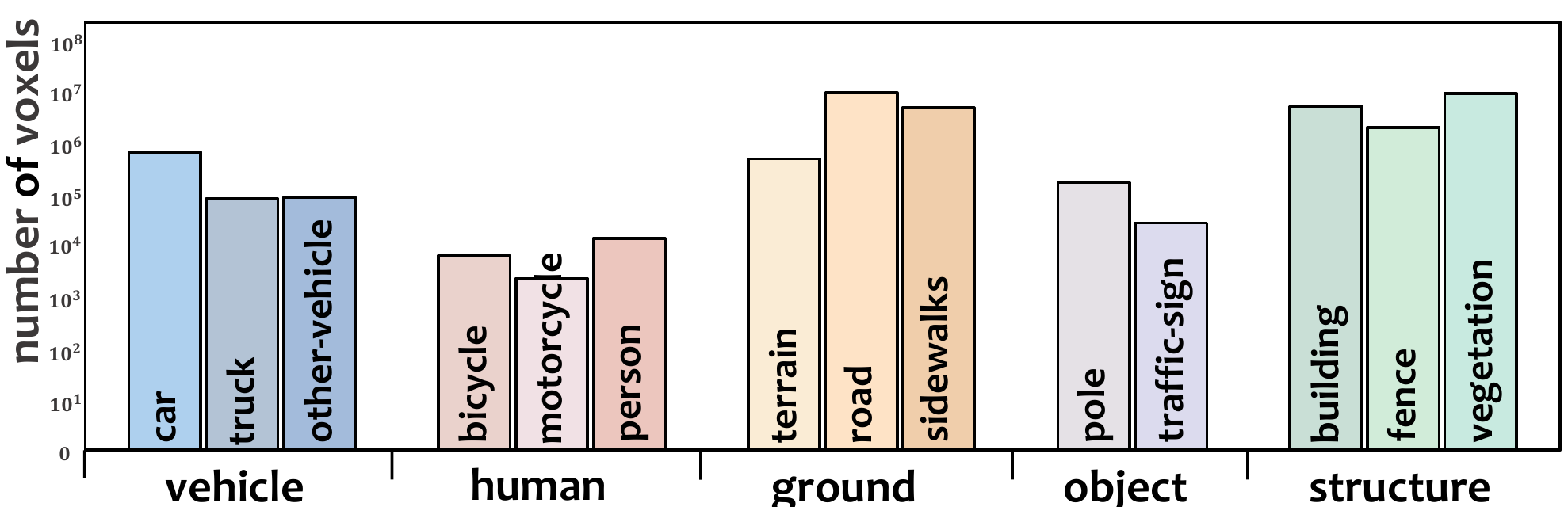}
\caption{\textbf{Overview of label distribution in the DSEC-SSC dataset.} The label distribution in the DSEC-SSC dataset is presented with the y-axis plotted on a logarithmic scale.
}
\label{fig:frequency}
\end{figure}

To address the lack of event-based 3D occupancy prediction datasets, we present DSEC-SSC, the first SSC dataset enhanced with real-world event modality, utilizing a streamlined 4D labeling pipeline that bypasses traditional reliance on 3D object detection or point cloud semantic annotations. 
Our approach includes three main stages: Semantic-guided Object Mapping, Spatiotemporal Purification of Dynamics, and Probability-guided Voxel Refinement, as illustrated in Fig.~\ref{fig:dsec_ssc_pipeline}.
DSEC-SSC includes $6$ sequences for training and $6$ for validation, consisting of $3,488$ frames with $14$ semantic classes, as shown in Fig.~\ref{fig:frequency}, which are grouped into five main categories: vehicle, human, ground, object, and structure. 
Each frame spans a region of $[{-}25.6m,{-}25.6m,{-}3m,25.6m,25.6m,3.4m]$ in 3D space with a voxel resolution of $0.4m$. Our proposed pipeline allows researchers to create their own 3D occupancy datasets with minimal manual effort, even with limited resources like Velodyne-16.

\subsection{Semantic-guided Object Mapping}
\label{sec:semantic_guided_dynamic_object_segmentation}
A primary challenge with DSEC~\cite{gehrig2021dsec} is the absence of point cloud semantic annotations, making it difficult to separate dynamic objects. Our 2D semantic-guided approach projects point clouds onto a 2D semantic map $\mathcal{I}_{\text{label}}$ provided by DSEC, enabling segmentation of static ($\mathcal{P}_{\text{Static}}$) and dynamic ($\mathcal{P}_{\text{Dynamic}}$) elements. This reduces manual annotation requirements, allowing efficient object isolation with only targeted human intervention.
After separating the dynamic and static point clouds, we obtain poses through a LiDAR-SLAM~\cite{xu2022fastlio2}, constructing both static maps $\mathcal{M}_{\text{S}}$ and dynamic maps $\mathcal{M}_{\text{D}}$.

\subsection{Spatiotemporal Purification of Dynamics}
\label{sec:dynamic_objects_4d_reconstruction}
Accurate and consistent labels are essential for 3D occupancy prediction. To obtain 4D ground truth for semantic scene completion and prevent dynamic objects from suddenly flickering between consecutive frames and trailing artifacts. Handling voxels for dynamic objects is extremely crucial,
which essentially arises from a lack of precise spatiotemporal positioning. 

\begin{figure}[t!]
    \centering
    \begin{subfigure}[b]{0.45\linewidth}
        \centering
        \includegraphics[width=\linewidth]{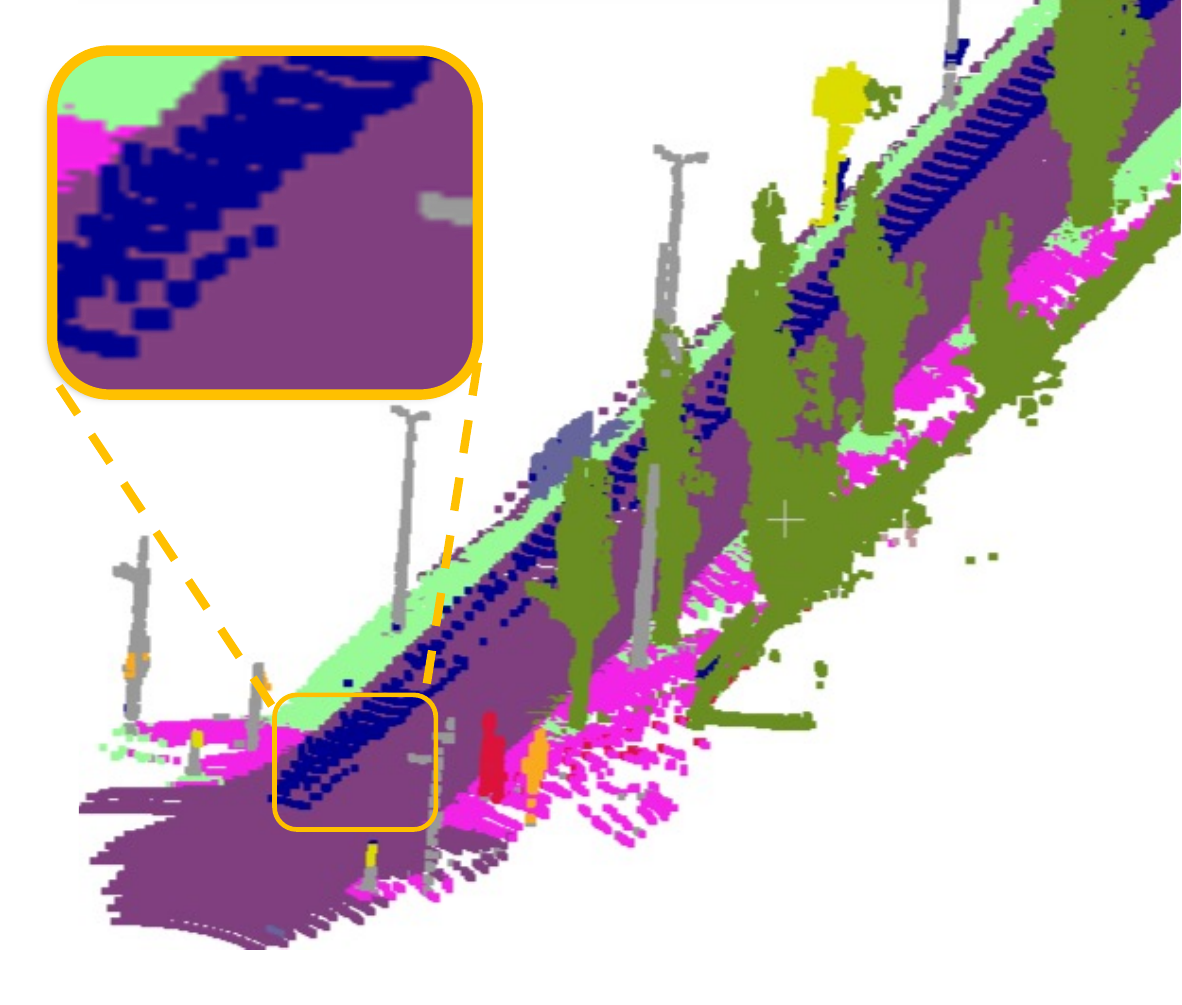}
        \caption{ }
        \label{fig:withoutDynamicObjectProcess}
    \end{subfigure}
    \hfill
    \begin{subfigure}[b]{0.45\linewidth}
        \centering
        \includegraphics[width=\linewidth]{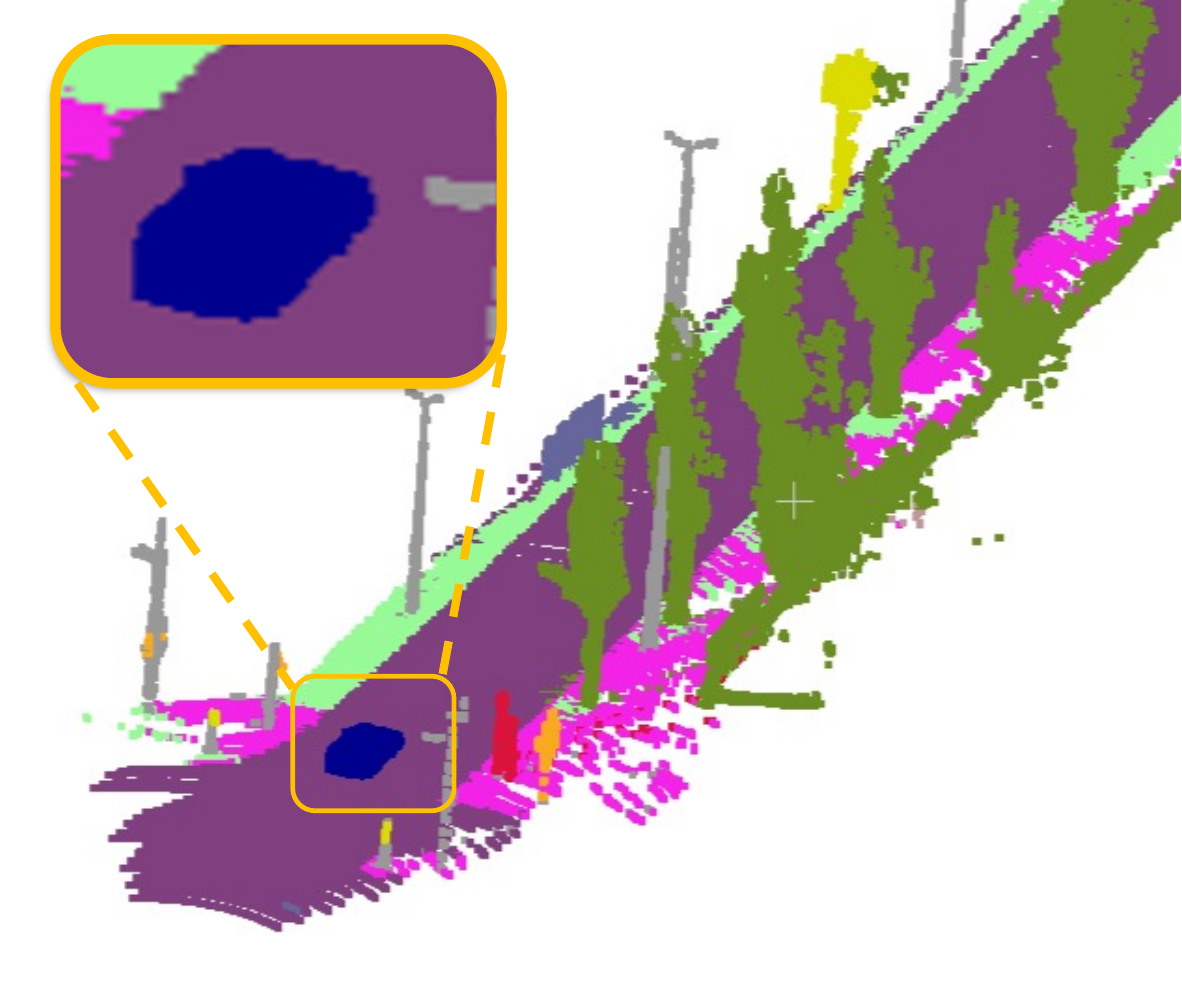}
        \caption{ }
        \label{fig:withDynamicObjectProcess}
    \end{subfigure}
    \caption{Comparison of point clouds (a) without and (b) with dynamic object processing.}
    \label{fig:car}
\end{figure}

Therefore, our approach to handling dynamic objects is based on two premises. First, DSEC uses the Velodyne-16 for mapping, resulting in very sparse captures of dynamic objects, making it difficult to map them completely and sometimes missing them altogether in certain frames. 
Second, we observed a ``pipeline'' effect in the 3D semantic map~\cite{roldao20223d}, which naturally retains spatiotemporal information. Leveraging this effect, we extract dynamic objects' positions and orientations from the pipeline map, then place reconstructed models of these dynamic objects at corresponding locations to achieve temporally continuous voxel representations.
To ensure dynamic objects are not lost when they are missed by LiDAR in certain frames, we incorporate manual supervision. First, we manually annotate ``pipeline'' instances in Bird’s-Eye-View (BEV). Next, we apply linear interpolation to estimate the position and orientation of any missing dynamic objects, providing reliable spatiotemporal information for dynamic objects $\mathcal{M}_{\text{D}}$. As shown in Fig.~\ref{fig:car}, this process effectively removes trailing artifacts from dynamic objects.

\subsection{Probability-guided Voxel Refinement}
\label{sec:probability_guided_voxel_refinement}
Labels obtained in the 2D semantic map are coarse. 
To acquire more accurate labels, the temporal information is leveraged. We propose a method based on Probability-guided Voxel Refinement, which can be divided into two steps: purification of the static scene through point cloud clustering and further refinement based on a voting mechanism.

The semantic labeling errors in the static point cloud stem from two main sources: one is the low accuracy of certain frames in the 2D semantic maps, and the other is occlusions in the 2D semantic map. Therefore, further purification in 3D is also required.
After performing plane fitting to remove the ground $\mathcal{M}_{\text{g}}$, we apply point cloud clustering to the non-ground points $\mathcal{M}_{\text{non-g}}$. 
For each cluster, we assign the semantic label with the highest occurrence probability as the semantic label for that cluster.
\begin{align}
&\mathcal{M}=\mathcal{M}_{\text{D}}+\mathcal{M}_{\text{S}},\quad\mathcal{M}_{\text{S}}=\mathcal{M}_{\text{g}}+\mathcal{M}_{\text{non-g}}, \label{eq:static} \\
&\mathcal{M}_{\text{g}}=\arg\max\sum_{i=1}^N \mathbf{1}(d_i<\epsilon),\ d_i=\operatorname{dist}\left({M}_{\text{S}},\mathcal{M}_{\text{g}}\right), \label{eq:ground} \\
&\mathcal{M}_{\text{non-g}}=\left\{\operatorname{K-means}\left(M_{\text{non-g},t}\right)\mid  t=1,2,\dots,N\right\}, \label{eq:non-ground} 
\end{align}
Here, \( \mathbf{1} \) is an indicator function, which counts a point as an inlier if the distance \( d_i \) is less than threshold \( \epsilon \), and \(M_{\text{S}}{\in}\mathcal{M}_{\text{S}}\).

Next, dynamic objects and static backgrounds are aggregated based on temporal information. During this process, we apply voxel-based voting within a bounding box defined by $[{-}25.6m,{-}25.6m,{-}3m,25.6m,25.6m,3.4m]$.
\begin{equation}
L_i = \arg\max_{l} \sum_{j} \delta(L_{i,j}, l).
\label{eq:Li}
\end{equation}
Here, \( V_i \) represents the \( i \){-}th voxel, and \( P_{i,j} \) denotes the \( j \){-}th point within \( V_i \), with each point \( P_{i,j} \) associated with a semantic label \( L_{i,j} \). 
The function \( \delta(L_{i,j}, l) \) is the Kronecker delta, which equals $1$ if \( L_{i,j} {=} l \) and $0$ otherwise. 
After the refinement operation, high-quality voxels are obtained.

\begin{figure*}[t!]
\centering
\begin{minipage}{1\linewidth}
    \centering
    \includegraphics[width=1\linewidth]{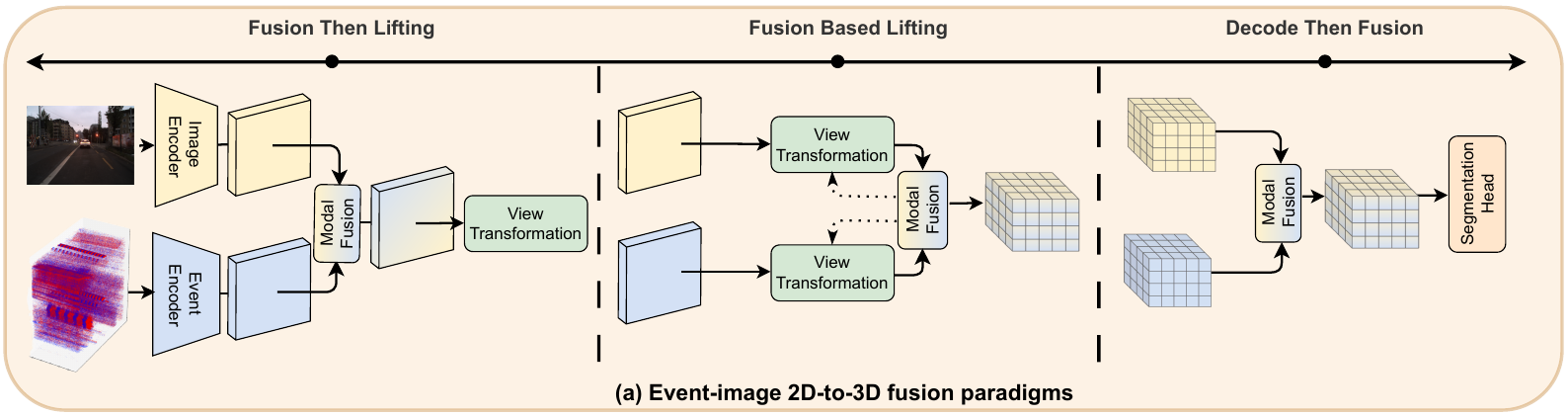}
\end{minipage}


\begin{minipage}{1\linewidth}
    \centering
    \includegraphics[width=1\linewidth]{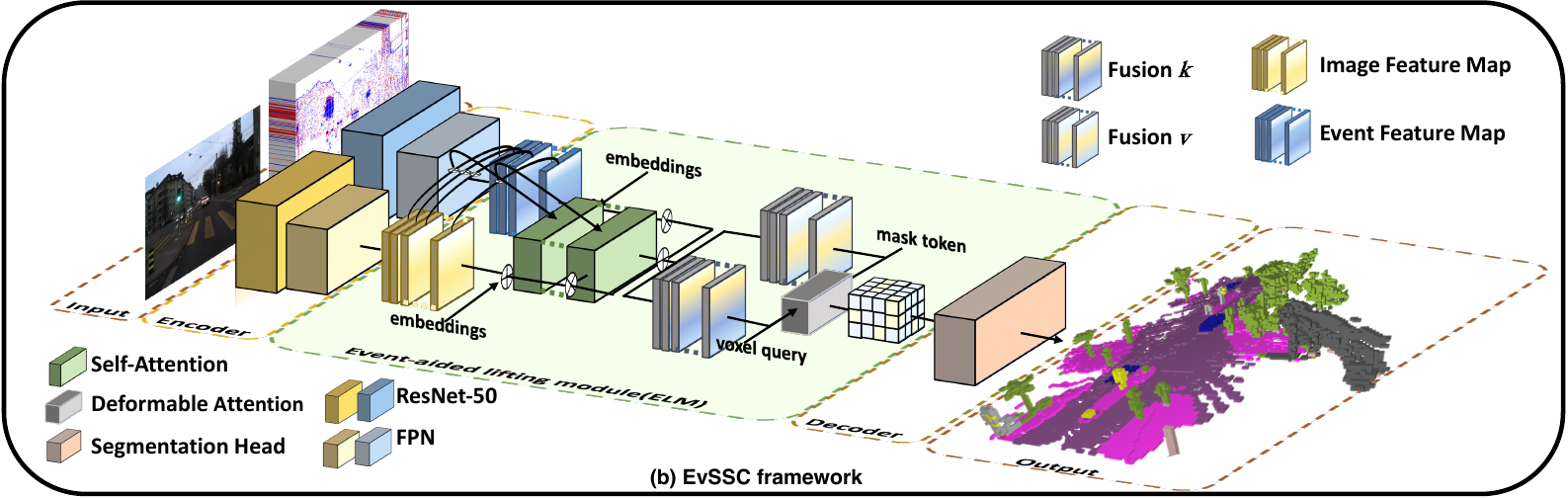}
\end{minipage}
\caption{\textbf{(a) Event-image 2D-to-3D fusion paradigms:} a spectrum of paradigms for feature fusion: (Left) performs 2D feature fusion after encoding, (Right) performs voxel fusion prior to the segmentation head. (Middle) ELM: fuse 2D or 3D features during the lifting process. \textbf{(b) EvSSC}: Given events and RGB images, 2D features are extracted. In ELM, by incorporating camera and level embeddings, the image \(k\)\&\(v\) and event \(k\)\&\(v\) are fused through self-attention to obtain the fusion \(k\)\&\(v\). Mask tokens and voxel queries are added to complete voxel features through deformable attention. 3D occupancy prediction is obtained through the segmentation head.}
\label{fig:fusion&framework}
\end{figure*}

%% file: Tex_content/method.tex
\section{EvSSC: Proposed Fusion Framework}
Current visual occupancy perception methods~\cite{li2023voxformer,mei2024sgn,cao2022monoscene} rely heavily on traditional cameras, limiting their robustness in dynamic scenes. 
To enable RGB-Event fusion-driven SSC, we develop \emph{EvSSC}, the first framework for event-aided semantic scene completion, introducing three fusion paradigms and proposing the \emph{Event-aided Lifting Module (ELM)}, designed for efficient integration of image and event features. ELM is flexible across various network architectures, making it adaptable to different SSC models.

\subsection{Fusion Paradigms}
The standard SSC network includes a camera encoder, 2D-to-3D transformation, 3D backbone, and completion head. The encoder extracts features through an image backbone and neck, which provide the foundation for constructing 3D volume features during the transformation stage, using either transformer-based~\cite{li2023voxformer,huang2023tpvformer} or LSS-based~\cite{mei2024sgn,philion2020lift,zhang2023occformer} methods. The 2D-to-3D lifting process is crucial in this pipeline, as it aligns 2D image features with the 3D spatial structure, enabling precise 3D occupancy predictions that are essential for autonomous perception. Accurate lifting ensures that spatial cues from 2D inputs are effectively translated into coherent 3D representations, directly impacting the model’s depth and spatial reasoning capabilities. To enable the effective fusion of image and event features within this framework, we propose three fusion paradigms, as shown in Fig.~\ref{fig:fusion&framework}.

\noindent\textbf{Fusion then lifting.} 
In this paradigm, image features \( \mathbf{F}^{\text{2D}}_{\text{img}} {\in} \mathbb{R}^{b \times c \times d} \) and event features \( \mathbf{F}^{\text{2D}}_{\text{event}} {\in} \mathbb{R}^{b \times c \times d} \) are fused at the 2D stage, where \( b {\times} c \) is the spatial resolution, and \( d \) is the feature dimension. $\oplus$ denotes the chosen fusion method:
\begin{align}
&\mathbf{F}^{\text{2D}}_{\text{fusion}} = \mathbf{F}^{\text{2D}}_{\text{img}} \oplus \mathbf{F}^{\text{2D}}_{\text{event}}.
\end{align}
This early fusion enhances feature alignment but may result in spatial inconsistencies in complex scenes.

\noindent\textbf{Fusion-based lifting.} 
Here, features for the image modality are represented as \( \mathbf{F}^{k}_{\text{img}} {\in} \mathbb{R}^{N_p \times d} \) and \( \mathbf{F}^{v}_{\text{img}} {\in} \mathbb{R}^{N_p \times d} \) for the key and value features, respectively. 
Similarly, the features for the event modality are represented as \( \mathbf{F}^{k}_{\text{event}} {\in} \mathbb{R}^{N_p \times d} \) and \( \mathbf{F}^{v}_{\text{event}} {\in} \mathbb{R}^{N_p \times d} \), where \( N_p \) is the number of query proposals, and \( d \) is the feature dimension.
The fused multi-modal features for key and value are then given by:
\begin{align}
&\mathbf{F}^{k}_{\text{fusion}} = \mathbf{F}^{k}_{\text{img}} \oplus \mathbf{F}^{k}_{\text{event}}, 
&\mathbf{F}^{v}_{\text{fusion}} = \mathbf{F}^{v}_{\text{img}} \oplus \mathbf{F}^{v}_{\text{event}}.
\end{align}
This approach combines the complementary features of both modalities, yielding robust fused representations for 3D volume construction.

\input{tables/Dsec_SSC}

\noindent\textbf{Decode then fusion.} 
In this paradigm, the fusion occurs after 2D-to-3D transformation. Features for the image modality are represented as \( \mathbf{F}^{\text{3D}}_{\text{img}} {\in} \mathbb{R}^{h \times w \times c \times d} \), and the features for the event modality are represented as \( \mathbf{F}^{\text{3D}}_{\text{event}} {\in} \mathbb{R}^{h \times w \times c \times d} \), where \( h {\times} w {\times} c \) denotes the 3D spatial resolution, and \( d \) is the feature dimension.
The fused multi-modal 3D features are then given by:
\begin{align}
&\mathbf{F}^{\text{3D}}_{\text{fusion}} = \mathbf{F}^{\text{3D}}_{\text{img}} \oplus \mathbf{F}^{\text{3D}}_{\text{event}}.
\end{align}
While this method preserves the independence of each modality’s features, it may delay cross-modal interaction, limiting early-stage fusion benefits.

\subsection{Event-aided Lifting Module}
The 2D-to-3D lifting process is pivotal in aligning 2D image features with 3D spatial structures, which is crucial for precise 3D occupancy predictions. 
Accurate lifting ensures effective translation of spatial cues from 2D inputs into coherent 3D representations, significantly impacting the model’s depth and spatial reasoning capabilities~\cite{philion2020lift,li2023bevdepth}. 
Fusion-Based Lifting offers potential benefits,
as it effectively combines complementary features from both modalities in the view transformation and enhances the robustness of 3D volume construction. 
Following this rationale, we design the \emph{Event-aided Lifting Module (ELM)} for adaptive 2D/3D feature fusion. 
In our ELM design, we combine key and value features from both image and event modalities in a self-attention framework. 
This integration allows the module to fuse spatial and temporal information adaptively.

Let the key and value features from the image encoder be represented as \( \mathbf{F}^{k}_{\text{img}} {\in} \mathbb{R}^{N \times d} \) and \( \mathbf{F}^{v}_{\text{img}} {\in} \mathbb{R}^{N \times d} \), respectively, and the corresponding features from the event encoder as \( \mathbf{F}^{k}_{\text{event}} {\in} \mathbb{R}^{N \times d} \) and \( \mathbf{F}^{v}_{\text{event}} {\in} \mathbb{R}^{N \times d} \), where \( N \) represents the number of spatial locations, and \( d \) is the feature dimension.
The image and event features are first added:
\begin{align}
   \mathbf{F}^{k}_{\text{add}} = \mathbf{F}^{k}_{\text{img}} \oplus \mathbf{F}^{k}_{\text{event}}, \quad \mathbf{F}^{v}_{\text{add}} = \mathbf{F}^{v}_{\text{img}} \oplus \mathbf{F}^{v}_{\text{event}}.
\end{align}
Within ELM, self-attention is applied to the aggregated features to generate the fused multi-modal representation. The attention scores are computed using the concatenated key and value features:
\vspace{-1em} 
\begin{equation}
   \text{Attention}(\mathbf{Q}, \mathbf{K}, \mathbf{V}) = \text{softmax} \left( \frac{\mathbf{Q} \cdot \mathbf{K}^T}{\sqrt{d}} \right) \cdot \mathbf{V},
\end{equation}

\begin{equation}
   w = \sigma\left( \text{G-Attention}(\mathbf{Q}, \mathbf{K}, \mathbf{V}) + \text{L-Attention}(\mathbf{Q}, \mathbf{K}, \mathbf{V}) \right),
\end{equation}

\begin{equation}
   \mathbf{F}^{k}_{\text{fusion}} = (1-w) \cdot \mathbf{F}^{k}_{\text{img}} + w \cdot \mathbf{F}^{k}_{\text{event}},
\end{equation}

\begin{equation}
   \mathbf{F}^{v}_{\text{fusion}} = (1-w) \cdot \mathbf{F}^{v}_{\text{img}} + w \cdot \mathbf{F}^{v}_{\text{event}},
\end{equation}
where \( \mathbf{Q} {=} \mathbf{F}^{k}_{\text{add}} \), \( \mathbf{K} {=} \mathbf{F}^{k}_{\text{add}} \), and \( \mathbf{V} {=} \mathbf{F}^{v}_{\text{add}} \).
After obtaining the fused multi-modal features from the self-attention module, deformable attention is applied for voxel querying, allowing the network to adaptively focus on relevant spatial information. Let \( \mathbf{Q}_{\text{voxel}} \) represent the voxel query features, and the deformable attention output is:
\begin{align}
   \mathbf{F}_{\text{voxel}} = \text{DeformableAttention}(\mathbf{Q}_{\text{voxel}}, \mathbf{F}^{v}_{\text{fusion}}).
\end{align}
The output \( \mathbf{F}_{\text{voxel}} \) represents the fused 3D voxel features, which are passed to the segmentation head for the final SSC.

%% file: tables/Dsec_SSC.tex
\begin{table*}[t]
    \centering
    \setlength{\tabcolsep}{0.0035\linewidth}
    \caption{\textbf{Semantic scene completion results using RGB and event data on the DSEC-SSC validation set}~\cite{gehrig2021dsec}\textbf{.}}
    \newcommand{\classfreq}[1]{{~\tiny(\dsecfreq{#1}\%)}}  %
    \centering
    \begin{tabular}{l|c|c|c c c c c c c c c c c c c c|c|c|c}
        \toprule
        Method & SSC Input & IoU
        & \rotatebox{90}{\textcolor{road_d}{$\blacksquare$} \makecell[l]{road \vspace{-3pt} \\ \classfreq{road}}} 
        & \rotatebox{90}{\textcolor{sidewalk_d}{$\blacksquare$} \makecell[l]{sidewalk \vspace{-3pt} \\ \classfreq{sidewalk}}}
        & \rotatebox{90}{\textcolor{building_d}{$\blacksquare$} \makecell[l]{building \vspace{-3pt} \\ \classfreq{building}}} 
        & \rotatebox{90}{\textcolor{car_d}{$\blacksquare$} \makecell[l]{car \vspace{-3pt} \\ \classfreq{car}}} 
        & \rotatebox{90}{\textcolor{truck_d}{$\blacksquare$} \makecell[l]{truck \vspace{-3pt} \\ \classfreq{truck}}} 
        & \rotatebox{90}{\textcolor{bicycle_d}{$\blacksquare$} \makecell[l]{bicycle \vspace{-3pt} \\ \classfreq{bicycle}}} 
        & \rotatebox{90}{\textcolor{motorcycle_d}{$\blacksquare$} \makecell[l]{motorcycle \vspace{-3pt} \\ \classfreq{motorcycle}}} 
        & \rotatebox{90}{\textcolor{other-vehicle_d}{$\blacksquare$} \makecell[l]{other-vehicle \vspace{-3pt} \\ \classfreq{othervehicle}}} 
        & \rotatebox{90}{\textcolor{vegetation_d}{$\blacksquare$} \makecell[l]{vegetation \vspace{-3pt} \\ \classfreq{vegetation}}} 
        & \rotatebox{90}{\textcolor{terrain_d}{$\blacksquare$} \makecell[l]{terrain \vspace{-3pt} \\ \classfreq{terrain}}} 
        & \rotatebox{90}{\textcolor{person_d}{$\blacksquare$} \makecell[l]{person \vspace{-3pt} \\ \classfreq{person}}} 
        & \rotatebox{90}{\textcolor{fence_d}{$\blacksquare$} \makecell[l]{fence \vspace{-3pt} \\ \classfreq{fence}}} 
        & \rotatebox{90}{\textcolor{pole_d}{$\blacksquare$} \makecell[l]{pole \vspace{-3pt} \\ \classfreq{pole}}} 
        & \rotatebox{90}{\textcolor{traffic-sign_d}{$\blacksquare$} \makecell[l]{traffic-sign \vspace{-3pt} \\ \classfreq{trafficsign}}}

        & mIoU & Precision & Recall \\
        \midrule\midrule
        
        \multicolumn{20}{c}{\textit{LiDAR-based}} \\ \midrule 
        SSCNet~\cite{song2017sscnet} & $x^{\text{lidar}}$ & 32.57 & 34.61 & 25.60 & 24.70 & 19.26 & 24.28 & 7.33 & 5.50 & 30.32 & 28.89 & 21.25 & 17.25 & 26.45 & 15.05 & 16.88 & 21.24 & 34.28 & \textbf{86.74}  \\ 
        SSCNet-full~\cite{song2017sscnet} & $x^{\text{lidar}}$ & 41.36 & 53.17 & 38.89 & \textbf{27.20} & 21.59 & 29.19 & 13.01 & 9.19 & 37.34 & \textbf{33.59} & 31.79 & \textbf{21.42} & 32.19 & \textbf{17.82} & 18.71 & 27.50 & 44.34 & 86.02  \\  
        LMSCNet~\cite{roldao2020lmscnet} & $x^{\text{lidar}}$ & 41.63 & 51.89 & 31.34 & 27.07 & 0.75 & 6.61 & 0.00 & 0.00 & 28.75 & 30.30 & 2.35 & 0.00 & 24.59 & 1.13 & 0.12 & 14.64 & \textbf{68.26} & 51.63  \\   
        \midrule
        \multicolumn{20}{c}{\textit{Vision-based}} \\ \midrule 
        \multirow{2}{*}{MonoScene~\cite{cao2022monoscene}} & $x^{\text{rgb}}$ & 35.34 & 50.68 & 31.93 & 14.29 & 16.27 & 14.15 & 6.25 & 1.77 & 25.5 & 20.83 & 21.63 & 7.76 & 22.56 & 7.01 & 7.45 & 17.72 & 53.08 & 51.41 \\  
                                   & $x^{\text{event}}$ & 33.35 & 48.57 & 29.36 & 13.47 & 14.06 & 8.93 & 3.01 & 2.32 & 24.92 & 18.31 & 20.68 & 6.17 & 19.35 & 5.42 & 7.44 & 15.86 & 51.13 & 48.96 \\ \midrule
        \multirow{2}{*}{OccFormer~\cite{zhang2023occformer}} & $x^{\text{rgb}}$ & 41.52 & 59.20 & \textbf{43.27} & 23.43 & \textbf{27.30} & 30.42 & 9.94 & 0.00 & 41.50 & 28.18 & \textbf{36.08} & 8.9 & \textbf{35.19} & 12.29 & 13.07 & 26.34 & 56.71 & 60.78 \\ 
                                   & $x^{\text{event}}$ & 37.88 & 56.15 & 39.35 & 18.33 & 22.33 & 25.13 & 4.36 & 0.00 & 37.87 & 23.48 & 32.76 & 18.53 & 29.86 & 8.14 & 4.82 & 22.93 & 54.19 & 55.73 \\ \midrule
        \multirow{2}{*}{Symphonies~\cite{jiang2024symphonize}} & $x^{\text{rgb}}$ & 41.42 & 55.42 & 37.92 & 18.33 & 21.68 & 24.02 & 13.95 & 4.82 & 33.31 & 29.42 & 28.98 & 15.74 & 30.64 & 11.80 & 19.96 & 24.71 & 62.46 & 55.15\\   
                                   & $x^{\text{event}}$ & 42.23 & 54.35 & 38.00 & 17.20 & 20.85 & 24.02 & 16.45 & 3.26 & 37.94 & 30.75 & 26.61 & 4.53 & 31.60 & 9.49 & 21.10 & 24.01 & 65.65 & 54.21 \\ \midrule
        \multirow{2}{*}{VoxFormer~\cite{li2023voxformer}} & $x^{\text{rgb}}$ & 47.25 & \textbf{60.08} & 42.29 & 25.80 & 21.95 & 22.53 & 13.42 & 5.18 & 36.34 & 32.36 & 29.72 & 12.79 & 31.25 & 10.08 & 14.87 & 25.62 & 65.19 & 63.19  \\   
                                   & $x^{\text{event}}$ & 46.40 & 53.33 & 30.32 & 20.77 & 11.44 & 7.15 & 3.65 & 1.52 & 22.45 & 27.97 & 10.91 & 3.54 & 20.45 & 5.06 & 6.60 & 16.08 & 64.70 & 62.13 \\ \midrule
        \multirow{2}{*}{SGN-S~\cite{mei2024sgn}} & $x^{\text{rgb}}$ & 43.70 & 57.11 & 41.34 & 23.19 & 24.49 & 31.21 & 18.09 & \textbf{14.30} & 39.64 & 31.46 & 33.33 & 20.11 & 33.59 & 15.16 & 23.82 & 29.06 & 65.61 & 56.69 \\   
                                   & $x^{\text{event}}$ & 40.44 & 51.55 & 33.68 & 19.07 & 19.25 & 17.81 & 14.58 & 8.83 & 31.96 & 27.73 & 24.28 & 11.51 & 23.64 & 11.76 & 19.7 & 22.53 & 62.32 & 53.53 \\ \midrule

    
       \rowcolor{gray!20}EvSSC (VoxFormer) & $x^{\text{event}}$,$x^{\text{rgb}}$ & \textbf{47.29} & 59.80 & 42.84 & 26.01 & 21.97 & 23.73 & 15.15 & 5.89 & 37.25 & 32.53 & 30.46 & 13.44 & 31.97 & 11.46 & 16.21 & 26.34 & 65.85 & 62.66  \\
        \rowcolor{gray!20}EvSSC (SGN-S) & $x^{\text{event}}$,$x^{\text{rgb}}$ & 43.99 & 56.27 & 41.86 & 22.8 & 25.09 & \textbf{31.9} & \textbf{20.77} & 14.16 & \textbf{43.36} & 31.23 & 32.07 & 19.05 & 34.36 & 14.58 & \textbf{26.2} & \textbf{29.55} & 65.89 & 56.97  \\ 
                        
        \bottomrule
    \end{tabular}\\
    \label{table:dsec_ssc}

\end{table*}

%% file: Tex_content/Exp.tex
\begin{figure*}[!t]
\includegraphics[width=1.0\linewidth]{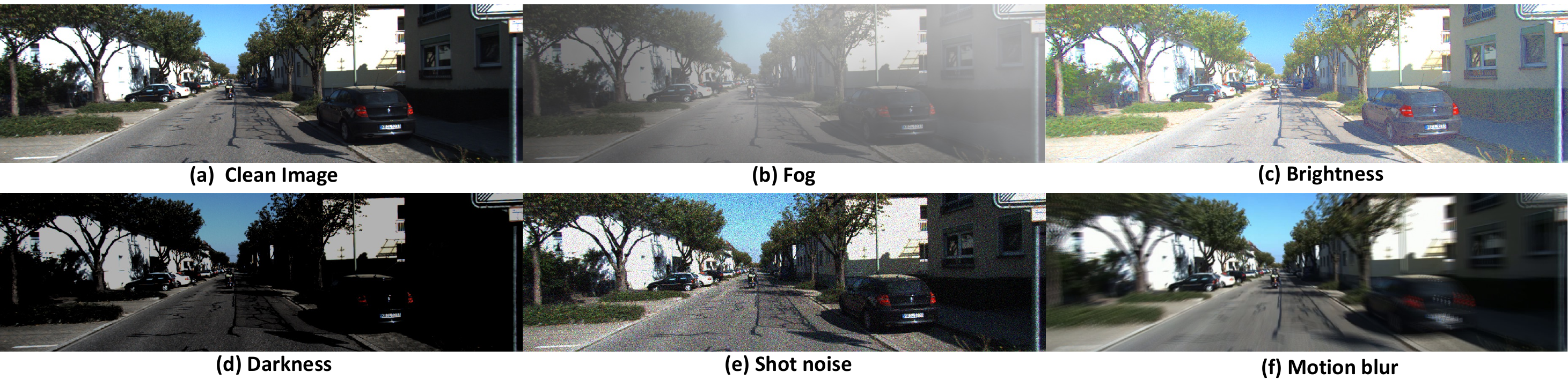}
\caption{\textbf{Overview of degradation modes in the SemanticKITTI dataset.} 
The degradation modes are illustrated, where clean image (a) serves as the baseline. Fog (b), brightness (c), darkness (d), and motion blur (f) represent degradations influenced by environmental factors or the interaction between the environment and sensors. Shot noise (e), primarily caused by sensor limitations or camera malfunctions, is included to evaluate the ability of event cameras to compensate for the shortcomings of traditional cameras in adverse scenarios.
}
\label{fig:corruption}
\end{figure*}
\input{tables/Kitti_SSC}

\section{Experiment}
\label{sec:exp}
\begin{figure*}[t]
\centering
\includegraphics[width=1\linewidth]
{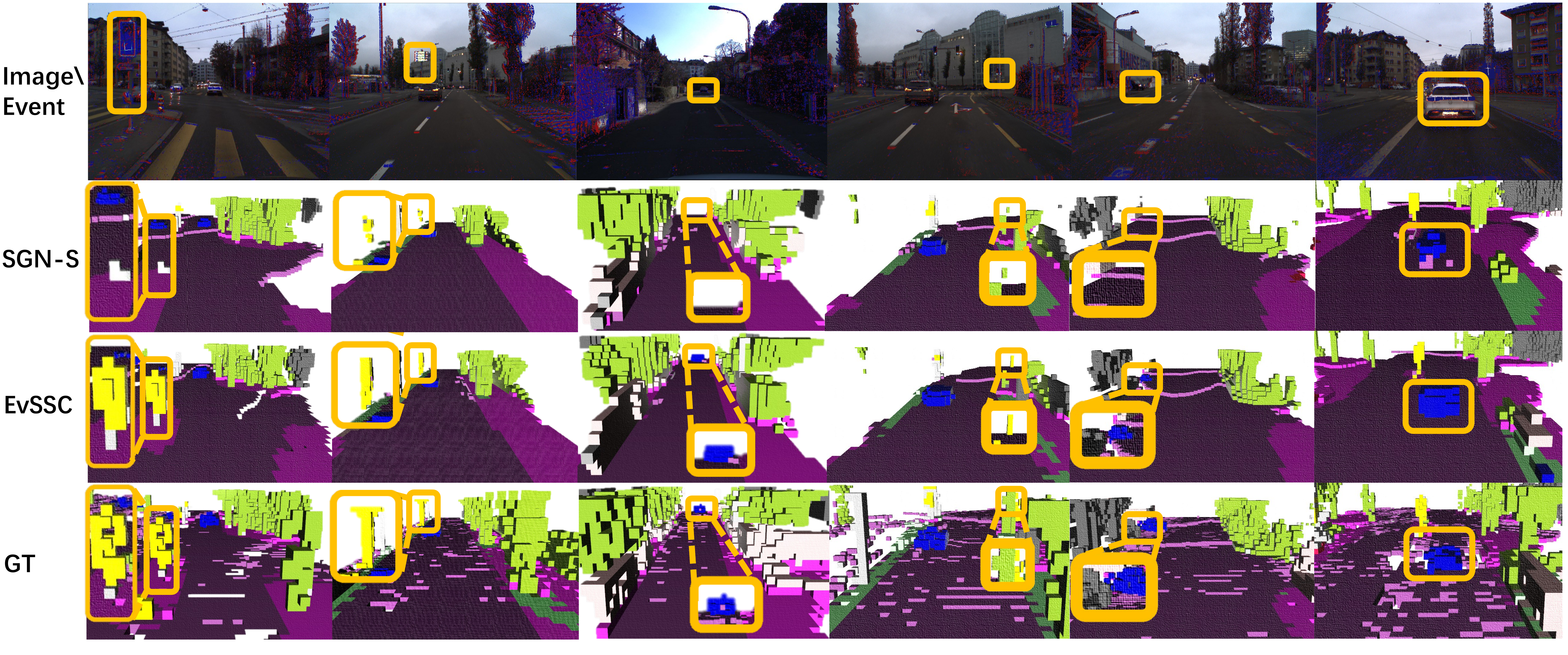}
\caption{\textbf{Qualitative results of EvSSC and baseline on DSEC-SSC.} EvSSC better captures scene layouts in low-light scenes.}
\label{fig:dsecquality}
\end{figure*}

\subsection{Datasets}
\noindent\textbf{DSEC}, introduced by Gehrig~\textit{et al.}~\cite{gehrig2021dsec}, consists of $24$ sequences captured in real-world outdoor driving scenes using event cameras, RGB cameras, and LiDAR. 
The dataset includes $7,800$ training samples and $2,100$ testing samples, captured at a size of $640{\times}480$. It covers both daytime and nighttime scenarios, with small and large motions. As the dataset does not provide official occupancy ground truth, we generated semantic occupancy labels by utilizing the released 2D semantic segmentation annotations and depth data, followed by manual refinement, resulting in the derived DSEC-SSC dataset (see Sec.~\ref{sec:benchmark} for details). 
LiDAR scans were voxelized into a $128{\times}128{\times}16$ grid with voxel sizes of $0.4m$, labeled into $14$ common categories (see Tab.~\ref{table:dsec_ssc}), consistent with semantic labels provided by DSEC.

\noindent \textbf{SemanticKITTI}~\cite{behley2019semantickitti} consists of outdoor LiDAR scans voxelized into a $256{\times}256{\times}32$ grid with $0.2m$ voxels, labeled into $21$ classes ($19$ \emph{semantic}, $1$ \emph{free}, $1$ \emph{unknown}). 
We use RGB images of size $370{\times}1,220$. We adopt the official $3834/815$ train/val splits and consistently evaluate at full scale (\textit{i.e.}, $1{:}1$). 
To assess the impact of the event modality on occupancy prediction for this dataset, we generate event sequences for SemanticKITTI using the official DVS-Voltmeter codebase~\cite{lin2022dvs}, resulting in the derived SemanticKITTI-E dataset. 
Additionally, to evaluate SSC's robustness against camera corruption, we simulate five common real-world corruptions including \emph{motion blur}, \emph{fog}, \emph{brightness}, \emph{darkness}, and \emph{shot noise}.

\subsection{Quantitative Comparison}
\noindent \textbf{Analyses on the DSEC-SSC benchmark.} 
We first benchmark popular camera occupancy prediction methods~\cite{cao2022monoscene,zhang2023occformer,li2023voxformer,mei2024sgn} on the newly proposed DSEC-SSC dataset. 
For a fair comparison, we report training results using both 2D images ($x^{rgb}$) and event data ($x^{event}$) as inputs. 
Note that we did not modify the baseline model structures. 
As a framework approach, EvSSC integrates both image and event modalities, making it applicable to various baselines. 
As shown in Tab.~\ref{table:dsec_ssc}, using only the event modality decreases mIoU and IoU across all camera methods. 
For example, the event modality version of MonoScene~\cite{cao2022monoscene} shows a drop in mIoU compared to the RGB modality version by $1.86$ ($15.86$ \textit{vs.} $17.72$), which is expected as the event modality lacks the detailed textures of RGB. 
EvSSC proves effective across different RGB SSC baselines. 
EvSSC (VoxFormer) improves mIoU by $0.72$ compared to RGB-only VoxFormer ($26.34$ \textit{vs.} $25.62$) and by $10.26$ compared to its event modality counterpart ($26.34$ \textit{vs.} $16.08$). Similarly, EvSSC (SGN) boots mIoU by $0.31$ compared to its RGB baseline ($29.37$ \textit{vs.} $29.06$) and by $6.84$ compared to its event modality counterpart ($29.37$ \textit{vs.} $22.53$).

\noindent \textbf{Analyses on the SemanticKITTI-E benchmark.}
As shown in Tab.~\ref{table:kitti_e_ssc}, we compare EvSSC, which fuses event and RGB modalities, with RGB-only occupancy prediction methods on SemanticKITTI-E. 
Despite the simulated nature of event data on this dataset, EvSSC effectively shows the complementary strengths of event and RGB modalities. 
EvSSC (VoxFormer) improves mIoU by $5.8\%$ over VoxFormer~\cite{li2023voxformer} ($13.61$ \textit{vs.} $12.86$). 
Using SGN-S~\cite{mei2024sgn} as the baseline, EvSSC (SGN) achieves the highest accuracy among visual occupancy models, reaching mIoU of $15.15$, surpassing 
Symphonies~\cite{jiang2024symphonize} ($15.15$ \textit{vs.} $14.89$). 
This further demonstrates the effectiveness of incorporating event modality to aid RGB in occupancy prediction, which comes with only a slight increase in terms of latency.
Evaluated on a single NVIDIA RTX 3090 GPU, the inference latencies of VoxFormer and EvSSC are $0.996s$ and $1.005s$ per frame, respectively.

\subsection{Enhancing Robustness with Event Modality}

Traditional visual occupancy models are highly susceptible to image degradation challenges, such as low-light conditions and motion blur. 
As shown in Tab.~\ref{table:dsec_ssc}, DSEC-SSC includes numerous low-light scenes, where we quantitatively demonstrate the robustness that the event modality brings to occupancy prediction in real-world settings. 
To further evaluate the impact of event modality under various degradation conditions, we generated the SemanticKITTI-C dataset, featuring five common degradation scenarios: \emph{motion blur}, \emph{fog}, \emph{brightness}, \emph{darkness}, and \emph{shot noise} from image sensors. 
Specifically, we consider two experimental setups, reporting occupancy mIoU for each.

\input{tables/Kitti_corrupted_combined}

1) \textbf{Out-Of-Domain (OOD):} 
In this setting, models are trained on clean data and tested directly on degraded scenes without prior exposure to degradation. 
As shown in Tab.~\ref{table:kitti_c_ssc}, both VoxFormer~\cite{li2023voxformer} and SGN~\cite{mei2024sgn} show large performance drops across all degradation types, especially in motion blur ($12.86{\rightarrow}8.54$), darkness ($12.86{\rightarrow}9.53$), and shot noise ($12.86{\rightarrow}8.10$). 
Although EvSSC is also trained only on clean data, the inclusion of the event modality effectively enhances the robustness of both baselines across all degradations. 
For example, with motion blur, EvSSC improves mIoU by $6.8\%$ ($9.12$ \textit{vs.} $8.54$) and $8.0\%$ ($8.91$ \textit{vs.} $8.25$) over the baselines, demonstrating the value of temporal cues in events; for darkness, EvSSC improves by $3.6\%$ ($9.87$ \textit{vs.} $9.53$) and $6.8\%$ ($9.78$ \textit{vs.} $9.16$), showing complementary strengths of the high dynamic range in event sensors.
\input{tables/Ablations}

2) \textbf{In-Domain (ID):} Here, models are pre-trained on degraded data. 
As shown in Tab.~\ref{table:kitti_c_ssc}, while overall occupancy prediction improves under ID training compared to OOD, accuracy remains lower than on clean data. 
Including events significantly enhances the in-domain performance of both baselines across all degradations. 
Especially in shot noise, EvSSC further improves by $52.5\%$ ($12.64$ \textit{vs.} $8.29$) and $5.1\%$ ($14.32$ \textit{vs.} $13.62$), confirming the robustness of event-RGB fusion when the image sensor partially fails.
\begin{figure*}[t!]
\centering
\begin{minipage}{1\linewidth}
    \centering
    \includegraphics[width=1\linewidth]{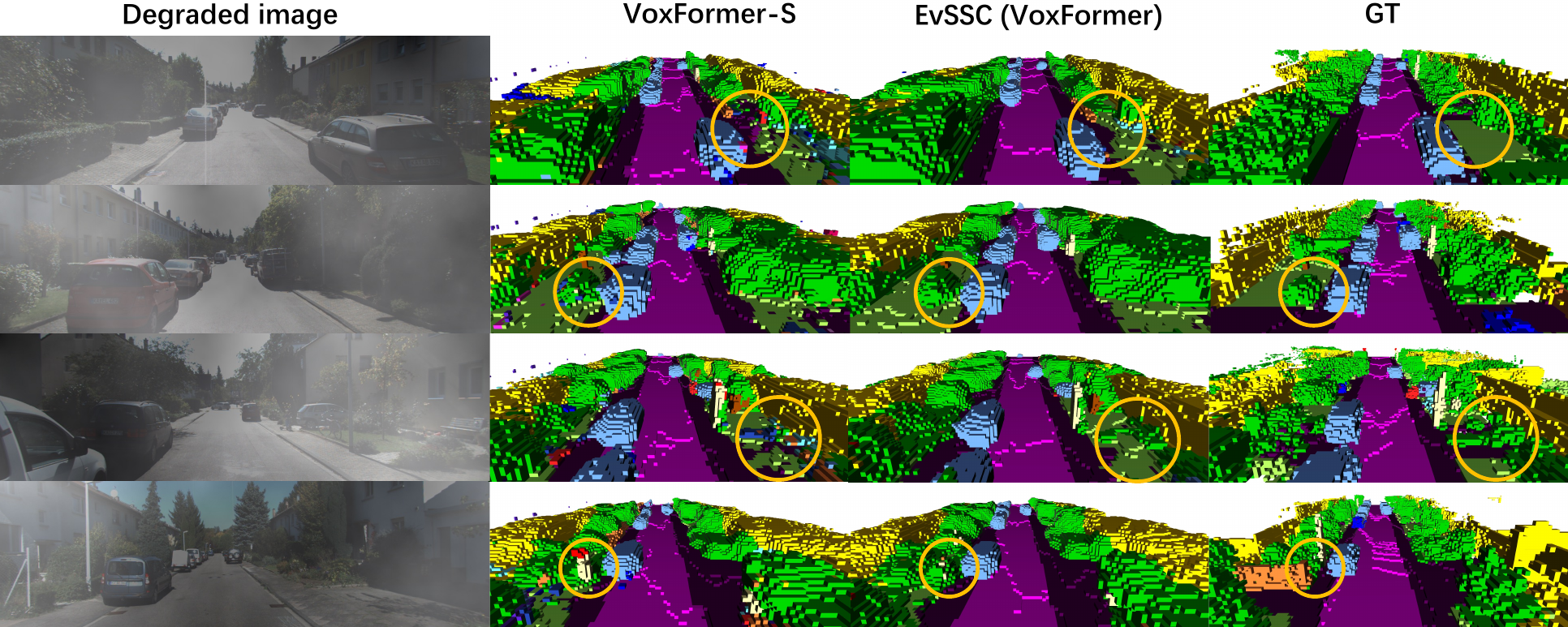}
\end{minipage}


\begin{minipage}{1\linewidth}
    \centering
    \includegraphics[width=1\linewidth]{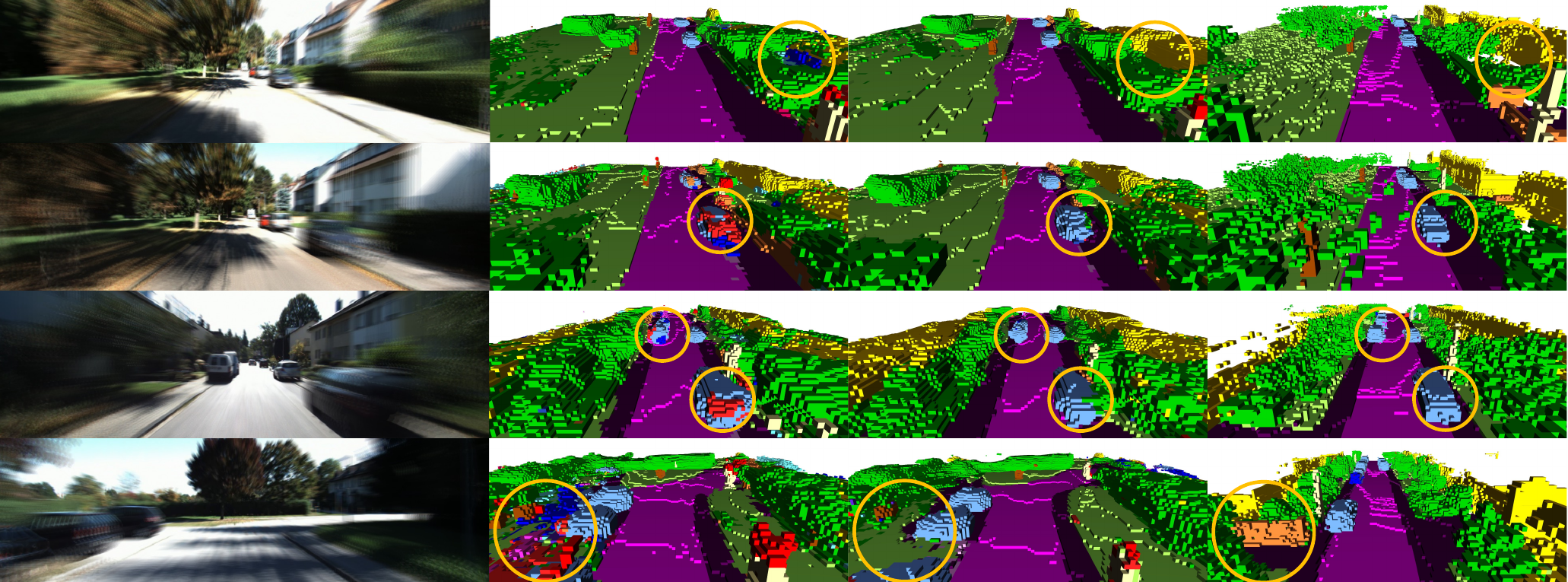}
\end{minipage}
\caption{\textbf{Qualitative results of EvSSC and baseline on SemanticKITTI-C.} More visual results of 3D occupancy on SemanticKITTI-C, showing predictions from VoxFormer-S~\cite{li2023voxformer} and EvSSC (VoxFormer). The first four rows correspond to the degradation mode of fog, while the last four rows correspond to the degradation mode of motion blur.}
\label{fig:corruption_v}
\end{figure*}
\subsection{Qualitative Comparison}
\noindent \textbf{Analyses on the DSEC-SSC benchmark.}
Based on the visualization results of EvSSC and the baseline SGN~\cite{mei2024sgn} on the DSEC-SSC validation set, as shown in Fig.~\ref{fig:dsecquality}, it can be observed that SGN often fails to fully reconstruct or detect traffic signs and cars in low-light conditions. 
In contrast, with the assistance of event data, EvSSC is capable of achieving complete reconstruction. Notably, in the fifth column, for vehicles with incomplete appearances in the image that SGN fails to detect, the integration of event data enables successful detection. 
Furthermore, as seen in the first and sixth columns, the incorporation of event data significantly improves the detection of lane markings and facilitates the reconstruction of detailed scene elements at the traffic intersections on low-contrast road surfaces.

\noindent \textbf{Analyses on the SemanticKITTI-C benchmark.}
As illustrated in Fig.~\ref{fig:corruption_v}, we visualize the prediction results of EvSSC and the baseline in scenarios with motion blur. It can be observed that due to the loss of texture information in the image, VoxFormer produces a significant number of mispredicted voxels. In contrast, with the assistance of event data, EvSSC achieves more accurate predictions, resulting in a cleaner and more refined visualization.

\subsection{Ablation Studies}
As shown in Tab.~\ref{tab:ablations}, we conduct ablations to validate design choices of EvSSC, reporting occupancy IoU and mIoU for various configurations.
(a) We examine the impact of three event-image fusion paradigms for 2D-to-3D occupancy models. 
To avoid bias introduced by complex fusion modules, we use simple additive fusion. 
Results show ``Fusion-Based Lifting'', which introduces event modality during the 2D-to-3D lifting process, provides the most significant mIoU gain ($13.10$ \textit{vs.} $12.86$).
Thus, EvSSC integrates the ELM module into the lifting stage.
(b) We explore the effects of different event representations on RGB-Event occupancy prediction.
The rasterized event representation used in this work encodes multiple statistics across channels, including counts and timestamps for positive and negative events. 
This approach effectively captures motion while aligning with RGB guidance, resulting in the highest mIoU gain over naive 2D event frames ($13.61$ \textit{vs.} $13.40$) and several representation methods~\cite{delbruck2008frame,vonmarcard2018recovering}.
(c) Building on (a), we investigate the impact of various attentional and RGB-X fusion methods~\cite{woo2018cbam,zhang2023cmx,sun2022event_deblurring} within a transformer-based lifting paradigm using VoxFormer~\cite{li2023voxformer}. 
Results show the dual-branch self-attention design in ELM provides the largest boost compared to simple addition ($13.61$ \textit{vs.} $13.10$).
(d) To assess the generalizability of ELM across architectures, we test ELM on LSS-style~\cite{philion2020lift} SGN~\cite{mei2024sgn}. 
Results confirm the effectiveness of dual-branch self-attention in this setup as well, with mIoU boosting from $14.58$ to $15.15$.

\subsection{Efficiency Analysis}
We further perform efficiency evaluations on the DSEC-SSC, SemanticKITTI-E, and SemanticKITTI-C datasets using a single NVIDIA RTX 3090 GPU. 
As shown in Tab.~\ref{table:efficiency}, our EvSSC consistently achieves great accuracy gains on all datasets and a remarkable $52.5\%$ improvement of mIoU in degraded scenarios while requiring only an additional $0.91$GB of GPU memory and a $0.9\%$ increase in latency.
Furthermore, on the non-degraded SemanticKITTI-E dataset, a mere millisecond-level latency is required to achieve a mIoU improvement of $0.75$. Efficiency analysis shows that minimal additional deployment costs can significantly improve prediction robustness, enhancing autonomous driving safety and practical application.

\input{tables/efficiency}

%% file: tables/Kitti_SSC.tex
\begin{table*}[h]
    \setlength{\tabcolsep}{0.0035\linewidth}
    \caption{\textbf{Semantic scene completion results on the SemanticKITTI validation set}~\cite{behley2019semantickitti}\textbf{.}}
    \newcommand{\classfreq}[1]{{~\tiny(\semkitfreq{#1}\%)}}  %
    \centering
    \begin{tabular}{l|c|c|c c c c c c c c c c c c c c c c c c c|c}
        \toprule
        Method & SSC Input & IoU
        & \rotatebox{90}{\textcolor{road_s}{$\blacksquare$} \makecell[l]{road \vspace{-3pt} \\ \classfreq{road}}} 
        & \rotatebox{90}{\textcolor{sidewalk_s}{$\blacksquare$} \makecell[l]{sidewalk \vspace{-3pt} \\ \classfreq{sidewalk}}}
        & \rotatebox{90}{\textcolor{parking_s}{$\blacksquare$} \makecell[l]{parking \vspace{-3pt} \\ \classfreq{parking}}} 
        & \rotatebox{90}{\textcolor{other-ground_s}{$\blacksquare$} \makecell[l]{other-ground \vspace{-3pt} \\ \classfreq{otherground}}} 
        & \rotatebox{90}{\textcolor{building_s}{$\blacksquare$} \makecell[l]{building \vspace{-3pt} \\ \classfreq{building}}} 
        & \rotatebox{90}{\textcolor{car_s}{$\blacksquare$} \makecell[l]{car \vspace{-3pt} \\ \classfreq{car}}} 
        & \rotatebox{90}{\textcolor{truck_s}{$\blacksquare$} \makecell[l]{truck \vspace{-3pt} \\ \classfreq{truck}}} 
        & \rotatebox{90}{\textcolor{bicycle_s}{$\blacksquare$} \makecell[l]{bicycle \vspace{-3pt} \\ \classfreq{bicycle}}} 
        & \rotatebox{90}{\textcolor{motorcycle_s}{$\blacksquare$} \makecell[l]{motorcycle \vspace{-3pt} \\ \classfreq{motorcycle}}} 
        & \rotatebox{90}{\textcolor{other-vehicle_s}{$\blacksquare$} \makecell[l]{other-vehicle \vspace{-3pt} \\ \classfreq{othervehicle}}} 
        & \rotatebox{90}{\textcolor{vegetation_s}{$\blacksquare$} \makecell[l]{vegetation \vspace{-3pt} \\ \classfreq{vegetation}}} 
        & \rotatebox{90}{\textcolor{trunk_s}{$\blacksquare$} \makecell[l]{trunk \vspace{-3pt} \\ \classfreq{trunk}}} 
        & \rotatebox{90}{\textcolor{terrain_s}{$\blacksquare$} \makecell[l]{terrain \vspace{-3pt} \\ \classfreq{terrain}}} 
        & \rotatebox{90}{\textcolor{person_s}{$\blacksquare$} \makecell[l]{person \vspace{-3pt} \\ \classfreq{person}}} 
        & \rotatebox{90}{\textcolor{bicyclist_s}{$\blacksquare$} \makecell[l]{bicyclist \vspace{-3pt} \\ \classfreq{bicyclist}}} 
        & \rotatebox{90}{\textcolor{motorcyclist_s}{$\blacksquare$} \makecell[l]{motorcyclist \vspace{-3pt} \\ \classfreq{motorcyclist}}} 
        & \rotatebox{90}{\textcolor{fence_s}{$\blacksquare$} \makecell[l]{fence \vspace{-3pt} \\ \classfreq{fence}}} 
        & \rotatebox{90}{\textcolor{pole_s}{$\blacksquare$} \makecell[l]{pole \vspace{-3pt} \\ \classfreq{pole}}} 
        & \rotatebox{90}{\textcolor{traffic-sign_s}{$\blacksquare$} \makecell[l]{traffic-sign \vspace{-3pt} \\ \classfreq{trafficsign}}} 
        & mIoU \\
        \midrule\midrule
        MonoScene~\cite{cao2022monoscene} & $x^{\text{rgb}}$ & 36.86 & 56.52 & 26.72 & 14.27 & 0.46 & 14.09 & 23.26 & 6.98 & 0.61 & 0.45 & 1.48 & 17.89 & 2.81 & 29.64 & 1.86 & 1.20 & 0.00 & 5.84 & 4.14 & 2.25 & 11.08 \\  
        OccFormer~\cite{zhang2023occformer} & $x^{\text{rgb}}$ & 36.63 & \textbf{59.45} & 28.10 & \textbf{21.44} & 0.33 & 11.27 & 15.09 & \textbf{25.42} & \textbf{9.91} & 2.21 & 1.52 & 19.40 & 3.53 & 31.99 & 3.50 & \textbf{3.87} & 0.00 & 5.96 & 4.03 & 2.52 & 13.13 \\ 
        VoxFormer~\cite{li2023voxformer} & $x^{\text{rgb}}$ & 44.42 & 57.20 & 28.68 & 13.66 & 0.36 & 19.12 & 27.37 & 5.22 & 0.40 & 0.53 & 4.12 & 25.83 & 6.24 & 33.29 & 1.11 & 1.58 & 0.00 & 7.66 & 7.53 & 4.46 & 12.86 \\  
        SGN~\cite{mei2024sgn} & $x^{\text{rgb}}$ & 43.60 & 59.32 & \textbf{30.51} & 18.46 & 0.42 & 21.43 & \textbf{31.88} & 13.18 & 0.58 & 0.17 & 5.68 & 25.98 & 7.43 & 34.42 & 1.28 & 1.49 & 0.00 & 9.66 & 9.83 & 4.71 & 14.55 \\
        Symphonies~\cite{jiang2024symphonize} & $x^{\text{rgb}}$ & 41.92 & 56.37 & 27.58 & 15.28 & \textbf{0.95} & 21.64 & 28.68 & 20.44 & 2.54 & \textbf{2.82} & \textbf{13.89} & 25.72 & 6.60 & 30.87 & \textbf{3.52} & 2.24 & 0.00 & 8.40 & 9.57 & 5.76 & 14.89 \\ \midrule
        \rowcolor{gray!20}EvSSC (VoxFormer) & $x^{\text{event}}$,$x^{\text{rgb}}$ & \textbf{45.01} & 57.68 & 28.74 & 16.43 & 0.48 & 19.43 & 27.51 & 11.27 & 0.59 & 0.78 & 5.05 & 25.95 & 6.82 & \textbf{34.92} & 1.61 & 1.96 & 0.00 & 7.68 & 7.46 & 4.21 & 13.61 \\
        \rowcolor{gray!20}EvSSC (SGN-S) & $x^{\text{event}}$,$x^{\text{rgb}}$ & 43.17 & 58.24 & 30.50 & 19.93 & 0.52 & \textbf{21.67} & 31.80 & 18.34 & 0.62 & 0.07 & 4.67 & \textbf{26.79} & \textbf{7.69} & 34.48 & 2.35 & 2.76 & 0.00 & \textbf{9.93} & \textbf{11.27} & \textbf{6.25}& \textbf{15.15} \\
        \bottomrule
    \end{tabular}\\
    \label{table:kitti_e_ssc}
\end{table*}

%% file: tables/Kitti_corrupted_combined.tex
\begin{table}[t!]
\centering
    \scriptsize
    \setlength{\tabcolsep}{0.0035\linewidth}
    \caption{\textbf{Semantic scene completion results on the corrupted SemanticKITTI-C dataset.}}
\setlength{\tabcolsep}{4pt} 
\resizebox{\columnwidth}{!}{%
\begin{tabular}{l|>{\columncolor{gray!10}}l>{\columncolor{blue!8}}l|>{\columncolor{gray!10}}l>{\columncolor{blue!8}}l}
\toprule
\textbf{Corruption} & \textbf{VoxFormer-S} & \textbf{EvSSC (VoxFormer)} & \textbf{SGN-S} & \textbf{EvSSC (SGN)}\\
\midrule\midrule
\multicolumn{5}{c}{\textit{Out-of-domain}} \\ \midrule 
\textbf{Motion Blur} & 8.54 & 9.12~\textcolor{blue}{ (+0.58)} & 8.25 & 8.91~\textcolor{blue}{ (+0.66)}\\ 
\textbf{Fog} & 8.28 & 9.13~\textcolor{blue}{ (+0.85)} & 8.35 & 8.90~\textcolor{blue}{ (+0.55)} \\ 
\textbf{Brightness} & 9.78 & 10.75~\textcolor{blue}{ (+0.97)} & 10.00 & 11.05~\textcolor{blue}{ (+1.05)} \\ 
\textbf{Darkness} & 9.53 & 9.87~\textcolor{blue}{ (+0.34)} & 9.16 & 9.78~\textcolor{blue}{ (+0.62)} \\
\textbf{Shot noise} & 8.10 & 8.83~\textcolor{blue}{ (+0.73)} & 5.87 & 6.07~\textcolor{blue}{ (+0.20)} \\
\midrule
\multicolumn{5}{c}{\textit{In-domain}} \\ \midrule 
\textbf{Motion Blur} & 11.57 & 12.35~\textcolor{blue}{ (+0.78)} & 13.02 & 13.55~\textcolor{blue}{ (+0.53)} \\ 
\textbf{Fog} & 12.38 & 13.13~\textcolor{blue}{ (+0.75)} & 14.23 & 14.55~\textcolor{blue}{ (+0.32)} \\ 
\textbf{Brightness} & 12.33 & 13.21~\textcolor{blue}{ (+0.88)} & 14.08 & 14.17~\textcolor{blue}{ (+0.09)} \\ 
\textbf{Darkness} & 11.48 & 11.84~\textcolor{blue}{ (+0.36)} & 13.16 & 13.29~\textcolor{blue}{ (+0.13)} \\
\textbf{Shot noise} & 8.29 & 12.64~\textcolor{blue}{ (+4.35)} & 13.62 & 14.32~\textcolor{blue}{ (+0.70)} \\
\bottomrule
\end{tabular}
}
\label{table:kitti_c_ssc}
\end{table}

%% file: tables/Ablations.tex
\begin{table}[t!]
    \scriptsize
    \centering
    \setlength{\tabcolsep}{0.0035\linewidth}
    \caption{\textbf{Ablation studies on fusion paradigms, RGB-Event fusion strategies, and event representation design on the SemanticKITTI-E dataset.}}
    \vskip-2ex
    \label{table:four_grid}
    \begin{subtable}[t]{0.48\linewidth}
        \caption{Ablations on fusion paradigms in EvSSC (VoxFormer-S).}
        \centering
        \begin{tabular}{p{2.5cm}|c|c}
            \toprule
            \rowcolor{gray!20}
            Aggregation & IoU & mIoU \\
            \midrule
            \midrule
            \textit{w/o} & 44.42 & 12.86 \\ \midrule
            Fusion then lifting & 45.05 & 12.86 \\
            \rowcolor{blue!8}\textbf{Fusion-based lifting} & 44.45 & \textbf{13.10}\\
            Decode then lifting & \textbf{45.54} & 12.68 \\  
            \bottomrule
        \end{tabular}
        \label{table:voxformer_add}
    \end{subtable}%
    \hfill
    \begin{subtable}[t]{0.48\linewidth}
        \caption{Ablations on event representation in EvSSC (VoxFormer-S). }
        \centering
        \begin{tabular}{p{2.5cm}|c|c}
            \toprule
            \rowcolor{gray!20}
            Representation & IoU
            & mIoU \\
            \midrule
            \midrule
            Timesurface~\cite{delbruck2008frame} & 44.92 & 13.49 \\  
            2D event frame & \textbf{45.09}  & 13.40 \\  
            HATS~\cite{vonmarcard2018recovering}& 44.91 & 13.44 \\          
            \rowcolor{blue!8}\textbf{Rasterized event} & 45.01 & \textbf{13.61}   \\
            \bottomrule
        \end{tabular}
        \label{table:voxformer_event_representation}
    \end{subtable}%

    \vspace{1mm} 

    \begin{subtable}[t]{0.48\linewidth}
        \caption{Ablations on RGB-Event fusion in EvSSC (VoxFormer-S).}
        \centering
        \begin{tabular}{l|c|c}
            \toprule
            \rowcolor{gray!20}
            Aggregation & IoU & mIoU \\
            \midrule
            \midrule
            \textit{w/o} & 44.42 & 12.86 \\ \midrule
            Add & 44.45 & 13.10 \\
            Concat & 44.51 & 13.08 \\
            CBAM~\cite{woo2018cbam} & 44.62 & 11.75 \\
            CMX (RGB-X)~\cite{zhang2023cmx} &44.41 & 12.58\\
            EFNet (RGB-Event)~\cite{sun2022event_deblurring} & 44.30 & 12.13 \\
            \rowcolor{blue!8}\textbf{ELM (ours)} & \textbf{45.01} & \textbf{13.61} \\ 
            \bottomrule
        \end{tabular}
        \label{table:voxformer_fbl}
    \end{subtable}%
    \hfill
    \begin{subtable}[t]{0.48\linewidth}
        \caption{Ablations on RGB-Event fusion in EvSSC (SGN-S).}
        \centering
        \begin{tabular}{l|c|c}
            \toprule
            \rowcolor{gray!20}
            Aggregation & IoU & mIoU \\
            \midrule
            \midrule
            \textit{w/o} & \textbf{44.60} & 14.55 \\ \midrule
            Add & 43.80 & 14.58 \\
            Concat & 43.41 & 14.66 \\ 
            CBAM~\cite{woo2018cbam} & 43.27 & 13.99 \\
            CMX (RGB-X)~\cite{zhang2023cmx} & 38.84 &11.52 \\
            EFNet (RGB-Event)~\cite{sun2022event_deblurring} & 43.63 & 14.64 \\
            \rowcolor{blue!8}\textbf{ELM (ours)} & 43.17 & \textbf{15.15} \\
            \bottomrule
        \end{tabular}
        \label{table:sgn_fbl}
    \end{subtable}%

\label{tab:ablations}

\end{table}

%% file: tables/efficiency.tex
\begin{table}[t!]
\centering
    \scriptsize
    \setlength{\tabcolsep}{0.0035\linewidth}
    \caption{\textbf{Computational efficiency of EvSSC across different datasets.} Memory denotes training memory usage.}
\setlength{\tabcolsep}{4pt} 
\resizebox{\columnwidth}{!}{%
\begin{tabular}{l|>{\columncolor{gray!10}}l>{\columncolor{blue!8}}l|>{\columncolor{gray!10}}l>{\columncolor{blue!8}}l}
\toprule
\textbf{ } & \textbf{VoxFormer-S} & \textbf{EvSSC (VoxFormer)} & \textbf{SGN-S} & \textbf{EvSSC (SGN)}\\
\midrule\midrule
\multicolumn{5}{c}{\textit{DSEC-SSC}} \\ \midrule 
\textbf{mIoU} &25.62 & 26.34 & 29.06 & 29.55\\ 
\textbf{IoU}  & 47.25 & 47.29 & 43.70 & 43.99\\ 
\textbf{Memory}  & 9.74G & 10.52G & 10.19G & 10.70G\\ 
\textbf{Latency} & 0.732s & 0.836s & 0.941s & 1.193s \\\midrule 
\multicolumn{5}{c}{\textit{SemanticKITTI-E}} \\ \midrule 
\textbf{mIoU} & 12.86 & 13.61 & 14.55 & 15.15\\ 
\textbf{IoU}  & 44.42 & 45.01 & 43.60 & 43.17\\ 
\textbf{Memory} & 14.87G & 15.78G & 15.29G & 17.79G \\ 
\textbf{Latency}  & 0.996s & 1.005s & 0.855s &1.005s\\ \midrule 
\multicolumn{5}{c}{\textit{SemanticKITTI-C Shot Noise}} \\ \midrule 
\textbf{mIoU} & 8.29 & 12.64 & 13.62 & 14.32\\ 
\textbf{IoU}  & 44.26 & 45.04 & 42.05 & 42.54\\ 
\textbf{Memory} & 14.87G & 15.78G & 15.29G & 17.79G \\ 
\textbf{Latency}  & 0.996s & 1.005s & 0.855s &1.005s\\
\bottomrule
\end{tabular}
}
\label{table:efficiency}
\end{table}

%% file: Tex_content/conclusion.tex
\section{Conclusion}
\label{sec:conclusion}

In this paper, We conduct the first comprehensive study on event-aided semantic scene completion. 
We present \emph{DSEC-SSC}, the first real-world benchmark for semantic scene completion incorporating event modality, primarily designed for visual perception in more challenging scenarios. 
Meanwhile, to enrich datasets,
we propose a deployable, general labeling pipeline that provides precise dynamic object processing. 
To enhance the robustness of visual perception, we introduce \emph{EvSSC}, including an \emph{Event-aided Lifting Module} for SSC. 
Extensive experiments demonstrate that EvSSC achieves significant performance gains on multiple benchmarks.

\noindent\textbf{Limitations and future work.} Although we have designed a highly deployable pipeline for semantic scene completion dataset construction and an efficient event-image fusion module, there are several ways to achieve further improvement:
\begin{itemize}
    \item \textbf{Robust LiDAR mapping framework:} In our future dataset construction, we will implement multi-sensor fusion by integrating GPS, IMU, and LiDAR SLAM to achieve enhanced trajectory and pose estimation accuracy, thereby establishing more reliable maps as the foundation for voxel generation.
    \item \textbf{Intelligent annotation automation:} To advance data annotation automation, we plan to develop AI-driven dynamic scene processors with uncertainty quantification, creating a human-in-the-loop framework to optimize annotation efficiency and reliability.
    \item \textbf{Temporal-aware event representation:} Rasterized events lose a small amount of time resolution. Our benchmark leaves space to further explore event representations and fusion modules that combine asynchronous events for better integration.
\end{itemize}